\documentclass{article}

\PassOptionsToPackage{numbers, compress}{natbib}

\usepackage[final]{neurips_2022}

\usepackage{subcaption}

\usepackage[utf8]{inputenc} % allow utf-8 input
\usepackage[T1]{fontenc}    % use 8-bit T1 fonts
\usepackage{hyperref}       % hyperlinks
\usepackage{url}            % simple URL typesetting
\usepackage{booktabs}       % professional-quality tables
\usepackage{amsfonts}       % blackboard math symbols
\usepackage{nicefrac}       % compact symbols for 1/2, etc.
\usepackage{microtype}      % microtypography
\usepackage{xcolor}         % 
\usepackage{caption}

\definecolor{CREL1}{HTML}{135ea0}
\definecolor{CREL2}{HTML}{b61651}
\definecolor{CREL3}{HTML}{b38600}

\definecolor{C1}{HTML}{1E88E5}
\definecolor{C2}{HTML}{D81B60}
\definecolor{C3}{HTML}{FFC107}
\definecolor{C4}{HTML}{004D40}
\definecolor{C5}{HTML}{D55E00}
\definecolor{C6}{HTML}{785EF0}

\usepackage{amsmath}
\usepackage{amssymb}
\usepackage{mathtools}
\usepackage{amsthm}
\usepackage{graphicx}
\usepackage{comment}
\usepackage{wrapfig}
\usepackage{booktabs} % for professional tables
\usepackage{makecell}       % compact symbols for 1/2, etc.

\usepackage[capitalize]{cleveref}

\title{Evaluation Beyond Task Performance:\\ Analyzing Concepts in AlphaZero in Hex}

\author{Charles Lovering\thanks{Equal contribution.} \quad Jessica Zosa Forde$^{*}$ \\
 \quad \textbf{George Konidaris} \quad
\textbf{Ellie Pavlick} \quad
\textbf{Michael L.\ Littman}\\
Department of Computer Science \\
Brown University\\
{\texttt{\{first\}}}\_{\texttt{\{last\}}}@brown.edu 
}

\begin{document}
\maketitle
\begin{abstract}
AlphaZero, an approach to reinforcement learning that couples neural networks and Monte Carlo tree search (MCTS), has produced state-of-the-art strategies for traditional board games like chess, Go, shogi, and Hex. While researchers and game commentators have suggested that AlphaZero uses concepts that humans consider important, it is unclear how these concepts are captured in the network. We investigate AlphaZero's internal representations in the game of Hex using two evaluation techniques from natural language processing (NLP): model probing and behavioral tests.
In doing so, we introduce new evaluation tools to the RL community, and illustrate how evaluations other than task performance can be used to provide a more complete picture of a model's strengths and weaknesses.
Our analyses in the game of Hex reveal interesting patterns and generate some testable hypotheses about how such models learn in general. For example, 
we find that MCTS discovers concepts before the neural network learns to encode them. We also find that concepts related to short-term end-game planning are best encoded in the final layers of the model, whereas concepts related to long-term planning are encoded in the middle layers of the model.\\

\href{https://bit.ly/alphatology}{\includegraphics[height=0.3cm]{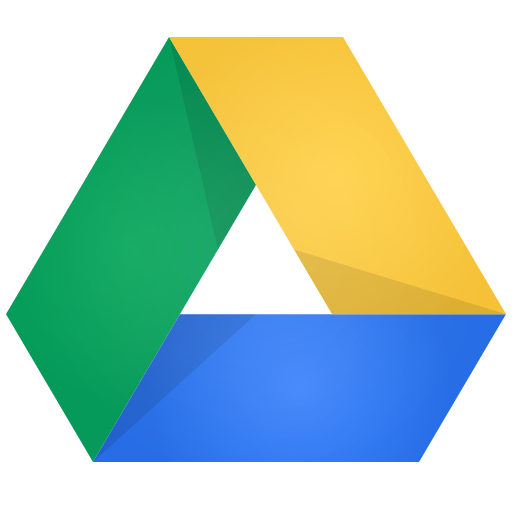}  https://bit.ly/alphatology}\hspace{0.25cm}
\href{https://github.com/jzf2101/alphatology}{\includegraphics[height=0.3cm]{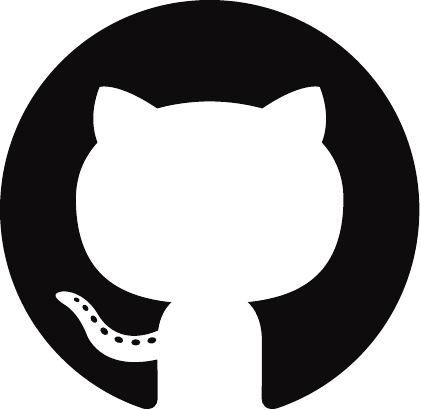}  https://github.com/jzf2101/alphatology}
\end{abstract}

\section{Introduction}\label{sec:introduction}

AlphaZero~\citep{Silver2016-ww}, a reinforcement-learning agent that combines Monte Carlo tree search~\citep{Brugmann1993-rw} with deep reinforcement learning, has achieved impressive performance at games like Go, chess, shogi, and Hex. Domain experts have observed that AlphaZero uses, but does not master, identifiable game concepts. For example, despite being exceptionally strong overall, AlphaZero appeared unable fully to project the implications of a \textit{ladder}---an important concept in the game of Go~\citep{Tian2019-kw}.  

Good performance can mask flaws in deep learning systems generally~\citep{Badgeley2019-yx,poliak2018hypothesis,gururangan2018annotation,Cooper2021-ul, Forde2021-sk, Zech2019-wj, Cooper2022-ml}, and deep reinforcement-learning systems in particular~\citep{Witty2018-sb,Zhang2018-ug}.  %\citep{poliak2018hypothesis,gururangan2018annotation}
Evaluating systems in terms of task performance alone makes it impossible to know if systems are ``right for the right reasons'' and difficult to predict how they will generalize to new situations~\citep{Witty2021-ge, Zech2019-oa}. Recently, the field of natural language processing (NLP) has begun to develop evaluation techniques that go beyond ``just'' task performance \citep{belinkov2022probing,belinkov2020interpretability}. For example, some techniques, \textit{probing classifiers}, inspect the form of models' internal representations to test whether they are consistent with linguistic theory \citep{Alain2017-xb,conneau-etal-2018-cram,tenney-etal-2019-bert}; other techniques, \textit{behavioral tests} or \textit{challenge sets}, evaluate specific types of out-of-distribution generalization to assess whether models encode human-like inductive biases \citep{ettinger2020bert,gauthier-etal:2020-syntaxgym, kim2019compositionality,lake2018generalization,linzen2016assessing, mccoy2020right,pandia2021sorting,warstadt2020blimp}. Although these techniques are only used over known concepts, often requiring expert domain knowledge to define, %; developing techniques to discover new concepts would be an exciting direction for future work. 
the insights generated by these techniques are not only scientifically interesting, but have begun to yield actionable insights on how to employ and improve models. For example, \citet{geva2020transformer} found that multilayer perceptrons (MLPs) in transformer layers act as key-value memories, and \citet{meng2022locating} leveraged this understanding to manipulate model behavior in a controlled manner. %

We demonstrate how these ideas leveraged in NLP can be applied to reinforcement learning, testing the capabilities and limitations of our models. Specifically, we leverage the above-described analysis techniques---probing classifiers and behavioral tests---to interpret AlphaZero's behavior at a conceptual level. We use probing classifiers to determine if reinforcement-learning agents encode tactical and strategic conceptual information. However, probing performance alone is insufficient: Information may be encoded but not used~\citep{lovering2021predicting}. To address this issue, we also use behavioral tests, which evaluate an agent's decisions in a situation tailored to require the understanding of a specific concept. 
%Using both probing classifiers and behavioral tests, we study how internal representations and external behavior relate.

 \begin{figure}
 \centering
%  \begin{subfigure}[t]{.3\textwidth}
    \begin{subfigure}[t]{.23\textwidth}
\centering
 \includegraphics[width=\linewidth]{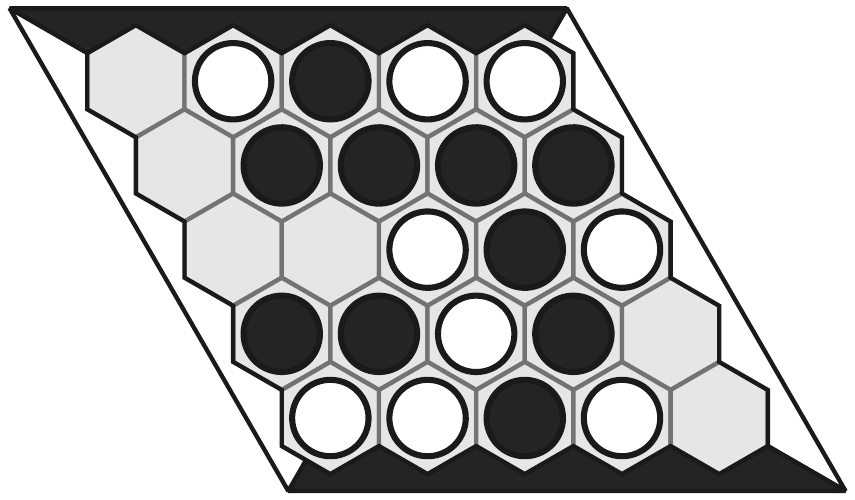}
   \caption{\textbf{An example winning board for black}, which connects the black edges together.}
   \label{fig:win}
 \end{subfigure}%
 \quad
 \begin{subfigure}[t]{.45\textwidth}
   \centering
 \includegraphics[width=\linewidth]{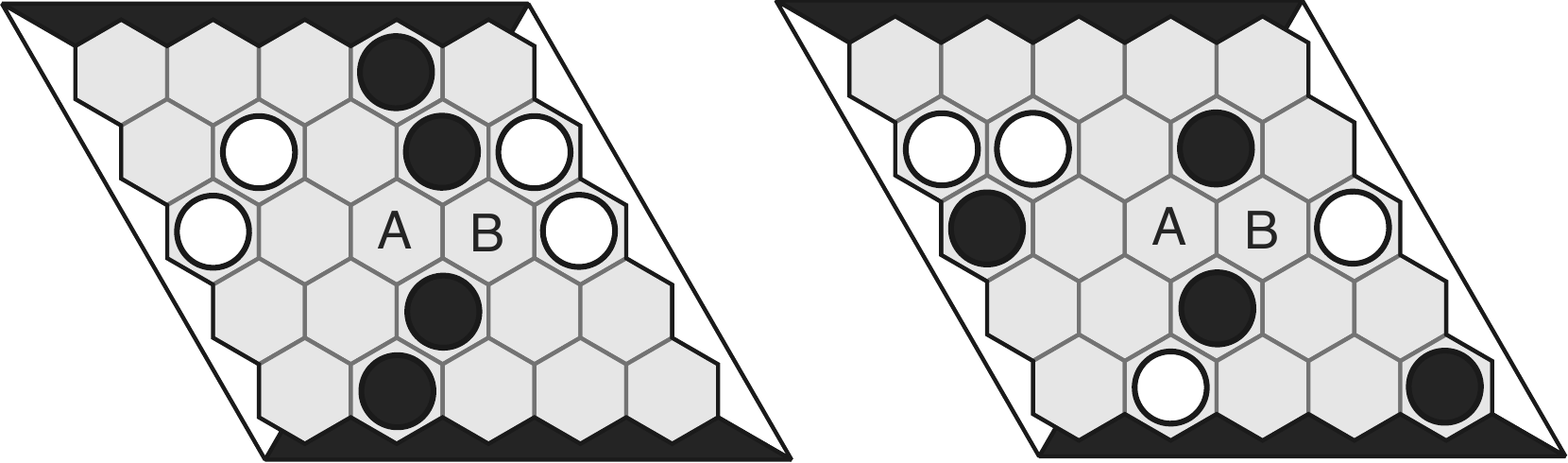}
   \caption{\textbf{Short- vs Long-term concepts.} Left: If black plays A or B, black immediately wins; the board concept containing A and B, \textit{bridge} (See \ref{sec:concepts}a), is relevant in the \textit{short-term}. Right: A and B can help black win only in the \textit{long-term}.}
   \label{fig:ex-short-long}
 \end{subfigure}
 \caption{To win in Hex, a player must use their pieces to form a connecting chain between the edges matching their color (a). Hex boards can be of varying size; we evaluate AlphaZero on a 9x9 board. In our work, we make a distinction between short- and long-term concepts (b). }
 \label{fig:ex-boards}
 \end{figure}
 
\begin{figure*}[h!]
    \centering
    \includegraphics[width=\linewidth]{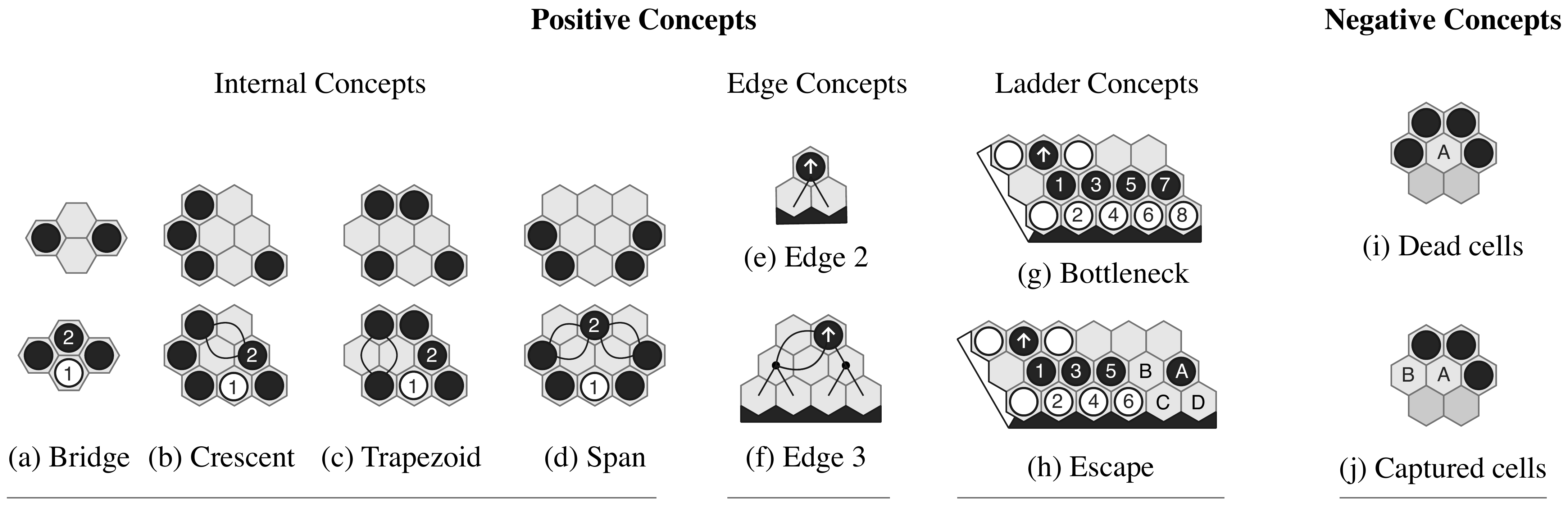} 
\caption{
\textbf{Hex templates exemplifying game concepts.} Concepts within the game of Hex are templates on the board formed by a player's pieces with known strategic and tactical implications. Positive concepts provide the player with the concept with multiple ways to connect the pieces within the concept together, despite possible attacks from the opponent.  An example of such a concept is the bridge (a). If white plays move 1, black can connect the two pieces of the bridge by playing move 2. Negative concepts change the strategic value of specific open spots of the board, such that the opponent is disincentivized to play those open spots, such as move A in (i). Each concept is further described in our Supplementary Material. Arrows indicate that the piece is connected to the opposite side of the board; the lines show the bridge concept within the other concepts.
}         \label{fig:concepts}
     \end{figure*}

Given these concept-level evaluation methods, we do an in-depth study of AlphaZero (AZ) trained to play Hex. Hex is a board game similar to Go (\S \ref{sec:concepts}), but provides an opportune test bed 
for analysis: the game is complex, yet has perfect-play baselines for smaller board sizes. Moreover, concepts and strategies within Hex have been studied by the Hex-playing community, making Hex a rich vector for study. Furthermore, the clear behavioral expectations in Hex make it easier to interpret and guide future work.\footnote{Again drawing inspiration from work in NLP, \citet{linzen-etal-2016-assessing} focused on one syntactic phenomenon in simple current neural networks, and the basic insight and methods enabled later studies to make broader claims about learning linguistic structure in many types of models~\citep{hu-etal-2020-systematic, warstadt2020blimp, warstadt-etal-2020-learning}.}
In our work, we probe for concepts that are taught to new players of Hex, like the \textit{bridge} (\cref{fig:concepts}a), and test that the model is able to use them to win games.
We investigate how concepts are represented within AZ, when concepts are learned during training, and where concepts are represented  within AZ's neural network.

Overall, our findings suggest that there are some regular dynamics to how AZ learns concepts with the game of Hex, and generate interesting predictions about how AZ behaves in general, which could be tested in other games. (In fact, concurrent work on chess \citep{mcgrath2021acquisition} and Go \citep{Tomlin2022} already begin to provide convergent evidence that some trends we see here generalize elsewhere.) We introduce a novel way of analyzing deep RL models on which such subsequent work can easily build. 

In summary, our main contributions are:
\begin{enumerate}

    \item We adapt several evaluation techniques from NLP to the RL setting, and illustrate how evaluations other than task performance can be used to provide a more complete picture of a model's strengths and weaknesses.
    \item We analyze a top-performing model from \citet{jones2021scaling}, and find that (1) short-term end-game concepts are best represented in the final layers of the network, whereas long-term concepts are best represented in the middle layers of the network; (2) concepts appear to originate with MCTS---with MCTS overriding the deep learning policy prediction early in training---but later in training these concepts are incorporated directly into the model's network.
    \end{enumerate}

\section{Concepts in Hex}\label{sec:concepts}
We study AlphaZero (AZ) agents trained to play Hex~\citep{Gardner1958-bp}, a game where two players take turns filling cells until one player builds a chain across the board. Hex has well-studied rules, reasonable computational costs, and it can be evaluated against perfect play, making it an ideal experimental vehicle for model probing. Unlike Go, there are no captures; once a cell is filled with a piece, the pieces stays there for the remainder of the game. For example, in \cref{fig:win}, white must connect pieces from left to right (marked on the edges), and black top to bottom.  Hex cannot end in a tie \citep{Gale1979-db}, and given perfect play, black, the first player, will win \citep{Gardner1958-bp}. \footnote{Hex is often played with a ``swap rule'' that makes the game more even between black and white. See \citet{jones2021scaling}, whose implementation we use, for further discussion on the swap rule. \citeauthor{jones2021scaling} did not include it to simplify the game implementation. }

In Hex, certain templates -- patterns of cells -- have been recognized as useful. Because building up a chain from one side of the board to the other is easily thwarted, a key part of learning how to play Hex is recognizing when it is possible to connect groups of pieces together. We consider these templates to be \textit{concepts} within the game. Concepts in Hex have different move implications depending on the condition of the board. We discuss two conditions, long- and short-term, in detail and provide additional discussion of conditions  in the Supplementary Material.

While the properties of concepts are debated \citep{margolis1999concepts}, here, in a board game setting, we consider a concept to be a useful template that generalizes across board configurations. For our purposes, ``understanding'' a concept amounts to recognizing it and leveraging its implications during gameplay (\S \ref{sec:concepts:taxonomy}).

\subsection{Long-term vs short-term Concepts}\label{sub:long-short}

We define a concept in Hex to be \textit{short-term} if its use is sufficient to win the game.  Concepts are typically short-term when there exists a connection between the concept and the player's board edges. For example, in \cref{fig:ex-short-long}, moves A and B belong to the \textit{bridge} concept (see Fig. \ref{fig:concepts}a).  In the left board, the board edges are connected, and all that is required for black to win is to play moves A or B.
Conversely, \textit{long-term} concepts are insufficient to winning the game if immediately used.  In the right board of \cref{fig:ex-short-long}, playing the \textit{bridge} concept is useful but insufficient for winning immediately; the concept is not connected to the edges of the board. Empirically, we find that whether the concept is short- or long-term has a significant impact on AlphaZero's representations (\cref{fig:short-long-summary}).

\subsection{Concept Taxonomy}
\label{sec:concepts:taxonomy}
From \citet{Seymour2019-hy} and \citet{King2004-fg}, we identify the nine concepts that we use in our analysis, summarized in \cref{fig:concepts}. These concepts fall into four categories, as described below.
\paragraph{Positive Concepts}
With the goal of Hex being to build a chain across the board, it is helpful to recognize when cells are virtually connected, that is, even in response to perfect adversarial play, the cells are guaranteed to connect \citep{hayward2005solving,pawlewicz2014stronger}. There are a few different types of positive concepts. All the positive concepts favor the player that owns the concept on the board. 

Internal concepts are templates that appear within the interior of the board. The \textit{bridge} (\cref{fig:concepts}(a)) is the simplest such concept. The larger internal templates -- \textit{crescent}, \textit{trapezoid}, \textit{span}  (\cref{fig:concepts}(b,c,d)) -- provide several possibilities to connect a player's pieces. \textit{Edge} concepts concern connecting a single cell to a given edge. Ladders in Hex are similar to ladders in Go. There are two different ladder concepts, \textit{bottleneck} (g) and \textit{escapes} (h). A  bottleneck favors the defender, because the attacking player cannot break through. An escape, however, allows the attacker to break through.

\paragraph{Negative Concepts}
Negative concepts do not inform one on which actions to play, but rather, which actions not to play. \textit{Dead cells} (\cref{fig:concepts}(i)) cannot impact the outcome of the game regardless of the color with which they are filled. While it is in general difficult to compute if a cell is dead \citep{bjornsson2006dead}, there are some known templates where it is easier to deduce. If a player can make a cell dead, that cell is \textit{captured}. Both captured and dead cells should not be filled.

\section{Designing Probing Tasks and Behavioral Tests for Hex} \label{sec:designing}
To understand the concepts encoded within AlphaZero, we probe its internal representations; to evaluate if these concepts are used by AZ, we test its behavior on tailored board configurations.
By evaluating AZ across training checkpoints, and across neural network layers, we can build up an understanding of how and where the model recognizes these concepts. Specifically, we evaluate the top-performing agent trained by \citet{jones2021scaling} across 21 training checkpoints. We additionally evaluate other publicly available agents \citep{jones2021scaling} from  that differ by width and depth and we report on those results in the Supplementary Material.  Code from \citeauthor{Jones2021-sb} is available under the MIT License.

\subsection{Reproducibility}\label{sub:reproducibility}
Our code and results are publicly available \citep{Lucic2022-fa}. Furthermore, we release example images of boards created for our probing classifiers and videos of the behavioral tests. The code, results and examples can be found at \url{https://bit.ly/alphatology}.  Our repository is also available on GitHub at \url{https://github.com/jzf2101/alphatology}. We report hyperparameters in the Supplementary Material. %\cref{app:tbl:hyper}. 
All error bars presented in plots are one standard deviation above and below the mean. We used  NVIDIA GeForce RTX 3090. The total compute across all experiments was about 24 GPU hours. We present a breakdown of the compute in the Supplementary Materials.

\subsection{Representational Probing}\label{sec:representation}

Model probing measures how well a model's learned representations encode a known concept \citep{Alain2017-xb}.
To make a model probe, one labels examples, such as Hex boards, with the presence or absence of a concept, such as the bridge concept (see Figure \cref{fig:concepts}). Next, one collects the model's activations for each example, and then trains a linear classifier (the \emph{probe}) to predict the presence of the concept based on activations. The linear classifier's test performance is used to interpret how well the original model encoded the concept.

We train a linear probe for each concept over each layer of AZ's network body. We follow a procedure similar to \citet{tenney-etal-2019-bert}: $H^{(0)}$ is the state of the board that is used as input into AZ.  For each board, we record the label $y_k$ to indicate the presence or absence of the concept $k$ on the board. $H^{(l)},$  $l \in 1..L$ is the activation of the layer $l$ of AZ's network body.\footnote{We never use activations from multiple layers at once, layer weights , nor AZ's value/policy outputs.} We then train linear classifiers $\mathcal{P}^{(l)}, l \in 0..L$ per layer to predict the presence vs.\ absence of a concept in a given board. These classifiers are our concept probes.

It is important to compare probing results against a baseline. We follow \citet{hewitt-liang-2019-designing}'s procedure to measure \textit{concept selectivity}, the delta between probing accuracy over a concept and random control. To form the random control, for each board  $H^{(0)}$ in the probing dataset, we  map each cell in that board to a random cell in a consistent manner to form the transposed board,  ${H}^{(0)}_{s}$. In this way, the same information is encoded in the original boards, but we expect the shuffled boards to be meaningless in Hex. Next, we train a set of linear probes ${\mathcal{P}}^{(l)}_{s}, l \in 0..L$ over the control boards ${H}^{(0)}_{s}$ to predict $y$. Now, finally, we can compute the concept selectivity by finding the delta in accuracy between ${\mathcal{P}}^{(l)}_s$ and $\mathcal{P}^{(l)}$. Concept selectivity is the performance of a probe beyond the performance of a probe on a control task, adding context for interpreting the results.

\paragraph{Implementation Details.}
Each concept is defined by a set of boards with vs.\ without that concept. We train and evaluate probing classifiers over AlphaZero's encoding of these boards. To generate a set of ($N = 2500$) boards for each concept, we translate the minimal templates across an empty board. Then, we add random enemy pieces to the board. Negative instances of a given concept match the statistics of the positive examples, except that the pieces pertaining to the concept template are randomly moved across the board.  This constitutes the long-term version of a concept. To form the short-term version of a concept, we connect the template to the edges of the board. 

\subsection{Behavioral Tests}\label{sec:behavioral}

Where model probing asks if concepts are represented within the model, behavioral tests asks if the model knows how to use the concept in gameplay. To interpret the behavioral tests, they must have clear behavioral expectations. We construct our behavioral tests for positive concepts (\S \ref{fig:concepts}) to be forced: If AZ understands the concept and plays the expected moves, AZ will win and pass the test; otherwise, AZ will lose the game and fail the test. For negative concepts (\S \ref{sec:concepts}), we have clear behavioral expectations. Dead and captured cells should never be filled. Thus, the behavioral test for negative cells checks that during a selfplay continuation of a board containing a dead (or captured) cell, the agent does not fill that cell.

Success on these behavioral tests are necessary but are alone insufficient to establish that the model has the concept. A negative result means that AZ is unable to use the concept, whereas a positive result means that AZ can use the concept to win games in forced situations. This approach has been used to test large language models. Specifically, we are inspired by Ettingers \citep{ettinger2020bert,pandia2021sorting} who uses simplified language to ask if the model is able to use a concept like negation or syntax when needed. If the model still fails to use the concept, then this good evidence that the concept is not represented.

\paragraph{Implementation Details.} 
For each concept, we create behavioral tests that comprise a board, forcing moves, and expected moves. (For the dead and captured cells there are no forcing or expected moves, only the moves to avoid). We discuss the motivation for our behavioral test setup and its connection to AlphaZero's style of gameplay in the Supplementary Material. By way of example, see \cref{fig:creating}. To generate a set ($N = 100$ samples) of boards, we translate the templates to a sampled valid board position. We then connect the concept to the edges of the board. Finally, we add connections for the defending player up to the region of the concept \textit{such that if the attacking player fails to complete the behavioral test, the other play would win}. Finally, we add the appropriate number of random pieces such that the board position is valid. The behavioral tests determine when/if AZ learns to navigate these situations correctly.

\begin{figure}[h!]
    \centering
    \includegraphics[width=0.55\linewidth]{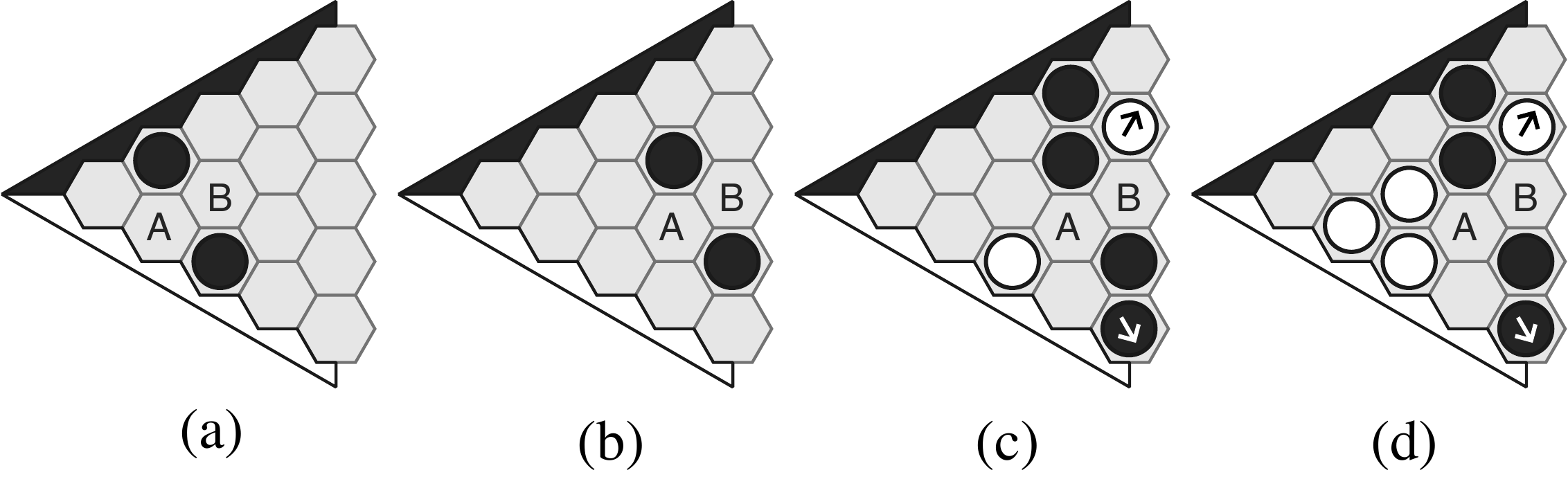}
\caption{\textbf{Creating behavioral tests from concept templates.} To evaluate AlphaZero's ability to utilize concepts during game play, we build synthetic boards where utilizing the strategic advantages of the given concepts allows the player to win the game. In this example, we demonstrate the building of a behavioral test for the bridge concept (Figure \ref{fig:concepts}a). 
The minimal template (a) is translated to a random position on the board (b). Then both players' pieces are connected to their respective edges they need to utilize to win the game (c). Finally, the minimum number of noise pieces necessary to form a valid board are added (d). Cells A, B are used to define the behavioral test. If white plays A, black must play B to win the game. (Only half a 5x5 board is shown for space.)
}
         \label{fig:creating}
     \end{figure}

\subsection{Considerations}
\label{sub:limitations}
\label{sub:impact}

\textbf{Limitations.} While we find a consistent relationship between the probing and behavioral tests in \cref{sec:discussion:ordering},  we do not run a counterfactual study. For example, we do not show that mistakes in recognizing a concept on a given board, lead to mistakes in using that concept. Studying the causal mechanisms of how the concept representations detected by model probing impact downstream model behavior is a rich direction for future work.

In our behavioral tests, all the concepts we tested share the property that they are about to be connected. (They share this property because we connect the concept so that the behavioral expectations are clear, \cref{fig:creating}.) All the concepts being tested could be interpreted as ``interrupt the opponent's soon-to-be-winning chain.'' However, the behavioral tests of different concepts report different performance levels (though admittedly similar), suggesting that the differences of concept are still relevant. Furthermore, the agent would still have had to use the targeted concepts as a constituent of ``interrupt the opponent's soon-to-be-winning chain.''

\textbf{Societal Impact.} Deep reinforcement learning models such as AZ are not interpretable \citep{Doshi-Velez2017-dp}, and yet are being applied to impactful, real world domains \citep{Gu2017-mq, Popova2018-jg, Gauci2018-lw, Bellemare2020-uu, jumper2021highly, Mirhoseini2021-ni}. In this work, we start to analyze how AZ comes to its decisions. Understanding a model's decisions is critical for accountability, but it does open the door to some avenues for exploitation. If it is known that a given algorithm does not reason about a concept well, this could be leveraged for ill. However, this risk only strengthens the argument for uncovering such problems and fixing them to prevent such risks.

\begin{figure}
\centering
\begin{subfigure}[t]{.6\textwidth}

    \centering
 \includegraphics[width=\linewidth]{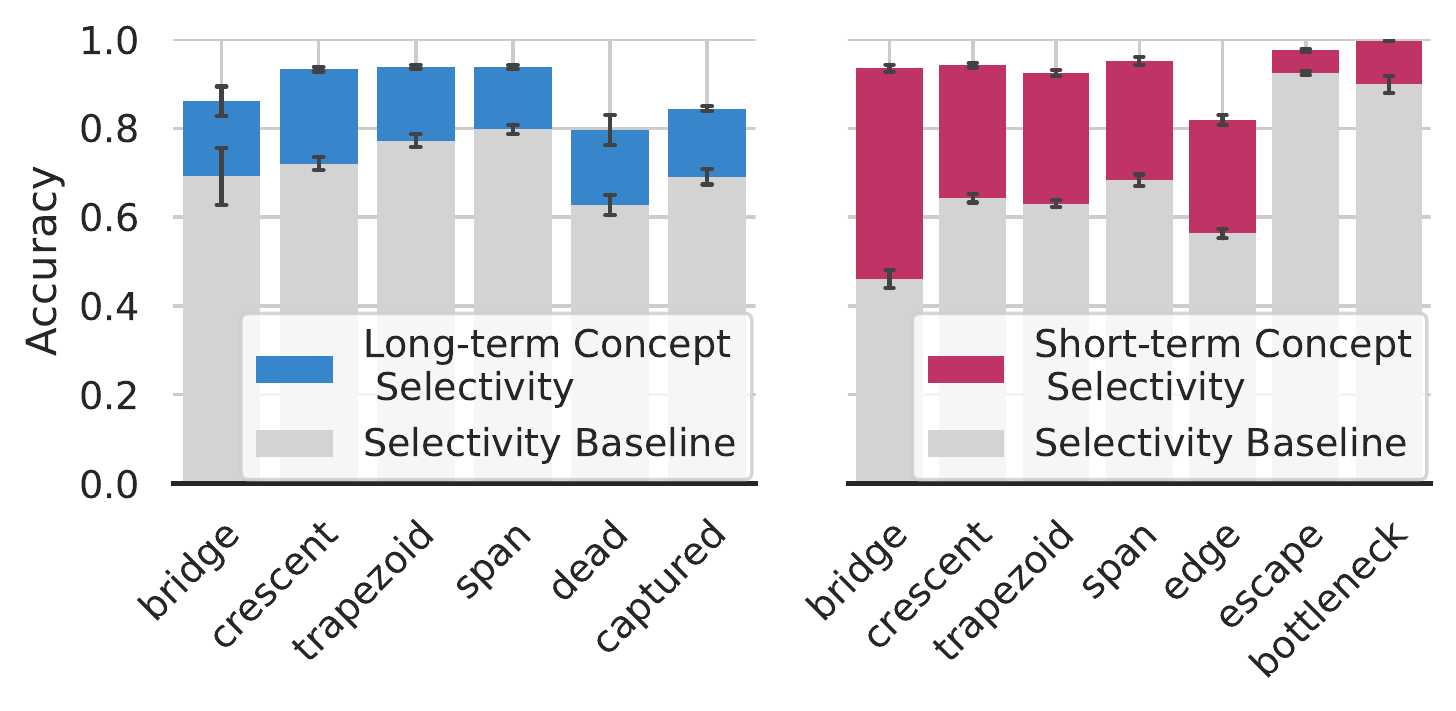}
\caption{\textbf{AlphaZero successfully encodes long-term and short-term concepts.} The selectivity \cite{hewitt-liang-2019-designing}, indicated by the colored bars, are the accuracy of a probe trained to identify a concept based on network activations, minus the accuracy of a selectivity baseline. The selectivity baseline randomly maps board pieces so that the new boards do not contain structures known to be relevant to Hex. We report selectivity based on the layer with the highest test accuracy.}
         \label{fig:selectivity}

\end{subfigure}%
\quad
\begin{subfigure}[t]{.355\textwidth}
  \centering
    \includegraphics[width=0.95\linewidth]{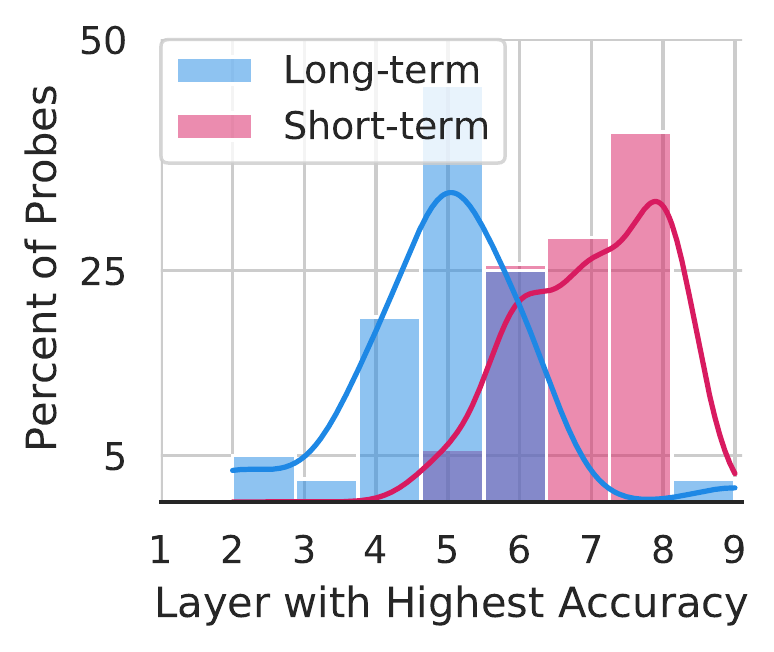}
    \caption{\textbf{Long-term concepts are best represented in the middle layers of the network whereas short-term concepts are best represented in the final layers of the network.} Each distribution shows 
    the layer in which probes had the highest accuracies.}
    \label{fig:best-layer}
\end{subfigure}
\caption{\textbf{Probing performance on long- and short-term concepts.}}
\label{fig:short-long-summary}
\end{figure}

\section{Results}\label{sec:results}
To understand which concepts AlphaZero (AZ) learns, we examine if its neural network activations encode the concepts (probing tests) and if AZ can use the concepts to win games (behavioral tests).

\subsection{AlphaZero Recognizes and Uses Concepts}
\begin{figure}[h!]
    \centering

    \includegraphics[width=\linewidth]{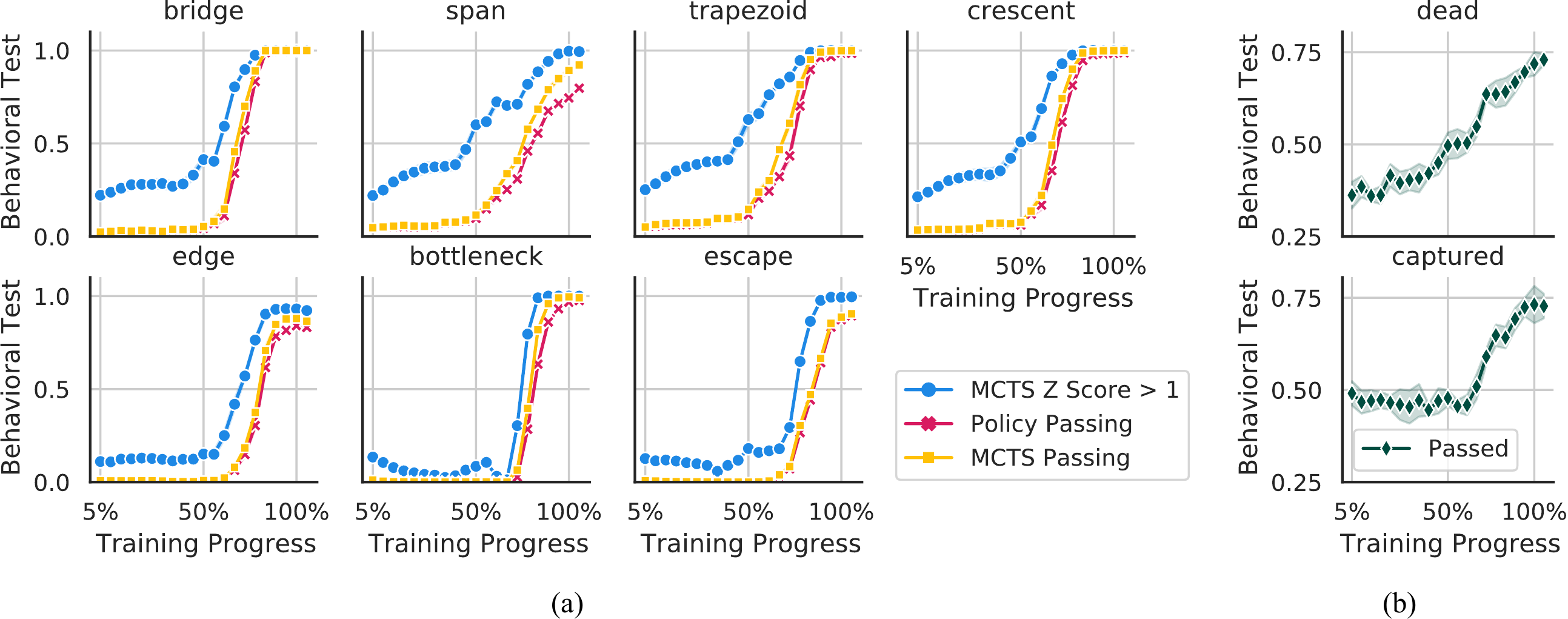}
\caption{\textbf{(a): AlphaZero learns to use positive concepts.}
At each checkpoint, we present AlphaZero (AZ) with a set of example boards that test its ability to utilize each concept. \textcolor{C3}{MCTS} and the \textcolor{C2}{policy network} both select actions that pass our behavioral tests with increasing frequency throughout training. We additionally report the rate at which the logit score of the action that passes our behavioral test is one standard deviation above the mean logit score \textcolor{C1}{(z score $> 1$)}.\\ \textbf{(b): AZ does not fully use the negative concepts.}
 \textcolor{C4}{Passed} denotes the rate at which AZ avoids the negative concept throughout selfplay rollouts. So, at the end of training, AZ plays moves in 25\% of our behavioral tests that will not impact the game \citep{bjornsson2006dead}.
}
         \label{fig:positive}
     \end{figure}

AZ successfully encodes short-term concepts, with high selectivity scores (\cref{fig:selectivity}). The long-term concept scores are also learned, but with slightly lower scores. By the end of training AZ is able to use all the positive concepts to win games (\cref{fig:positive}a). 

AZ improves on the behavioral tests 50\% of the way through training. Unsurprisingly, the MCTS passing rates increase before the  policy network passing rates (\cref{fig:positive}), though it need not have been the case. It is possible for AZ to represent the concepts before MCTS used them -- possibly via the signal through the value prediction.

Interestingly, the relative magnitude of the correct action logits initially increases earlier. To specify how we measure this, we need to cover two definitions. First, the logits are the outputs of the modules (MCTS or policy prediction head). Second, we use a Z-score, which reports the number of standard deviations higher a given value is than the mean of the population. In our case, the blue line in \cref{fig:positive}a, reports the proportion of cases where the Z-score of the correct action is greater than 1. So, this value captures when the correct action becomes more likely throughout training.

The trend in \cref{fig:positive}a suggests ``pre-conceptual'' information is learned, and coalesces (for bridge) 60\% of the way through training into an actionable understanding of the concept. (In \cref{sec:discussion:board}, we investigate further and find that this ``pre-conceptual'' information is not board structure.)

AZ also improves behaviorally upon negative concepts. However, it does not reach a perfect passing rate (\cref{fig:positive}b). The probing performance for the negative concepts, shown in \cref{fig:selectivity}, is also lower than for other concepts. This aligns with evidence that AZ wastes moves at the end of the game. This highlights a weakness in AlphaZero and a risk: Some concepts may be ``provable'' and useful to people, but ``deemed'' less important by AZ -- an agent that plays remarkably well. We will have to consider this dynamic as people begin to try to learn concepts from machines.

\subsection{Long-term and Short-term Concepts are Best Represented in Different Layers}
 \cref{fig:best-layer} highlight that short and long-term concepts are best represented at different layers. Long-term concepts, by the end of training, are best represented in the middle layers of the network. Short-term concepts, throughout training, are best represented in the upper layers of the network. We have two conjectures, which are not mutually exclusive, for why this may be the case. We leave the verification/refutation to future work. (1) Short-term concepts generally require more global information and so require more depth; (2) Short-term concepts factor directly into action selection and so are more proximal to the final task-specific layer of the network.

 We find that two and four layer networks demonstrate the same pattern; see our Supplementary Material. Lastly, \citet{mcgrath2021acquisition}'s results in chess are similar, although they did not directly study short- vs long-term concepts. The following is our interpretation of the results shown in their plots: ``in check'' (short-term) was best represented higher in the network and ``material imbalance'' (long-term) was largely represented lower in the network. ({Informally categorized, the short-term concepts (\citeauthor{mcgrath2021acquisition}'s Fig. 2)[c,e,f] are better represented in higher layers than are the long-term concepts (Fig. 2)[a,g,h,i].})

\subsection{Improvements in Behavioral Tests Occur Before Improvements in Probing Accuracy}\label{sec:discussion:ordering}
\begin{figure}[h]
    \centering
    \includegraphics[width=\linewidth]{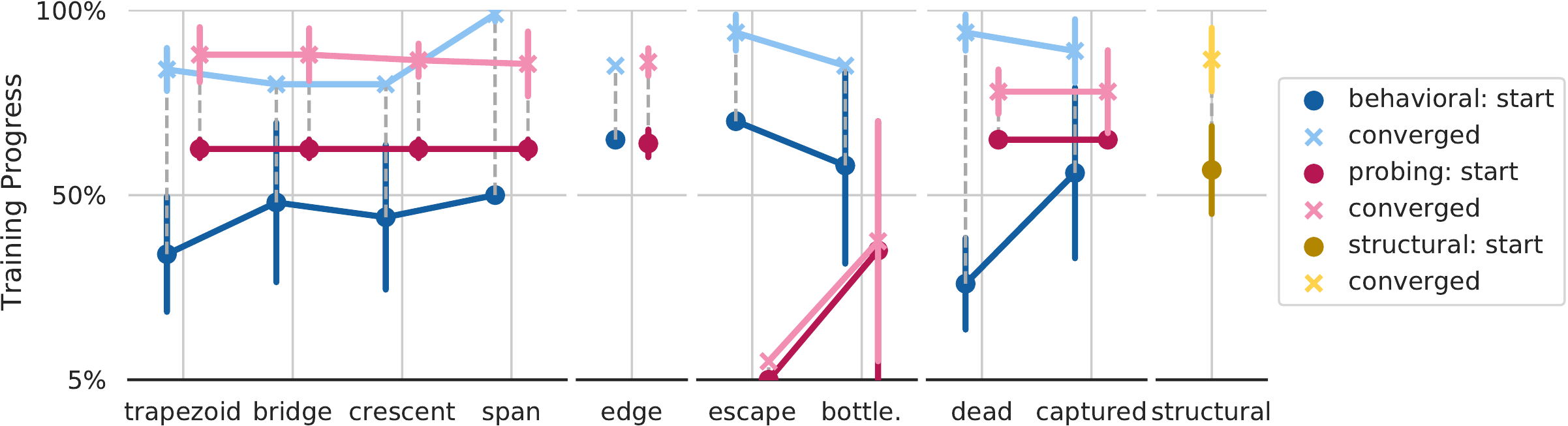}
    \caption{\textbf{Improvements in behavioral tests occur before improvements in probing accuracy.} Each point marks the mean checkpoint in training that AZ started to learn (or converge upon) the behavioral (or probing) test. Error bars are standard deviation across random seeds and short- vs long-term. We group the concepts as in \cref{fig:concepts}. While \textcolor{CREL1}{behavioral} tests start to improve before the   \textcolor{CREL2}{probing}, they both converge near the end of training. Exception: the ladder escape and bottleneck concepts (Fig. \ref{fig:concepts}h) are easy for the probes to detect, consequently having low selectivity (\cref{fig:selectivity}), perhaps because they occur along the edges of the board, and as a result, have fewer possible configurations.
    The concept \textcolor{CREL3}{structural}, (\S \ref{sec:discussion:board}), which evaluates how well AZ's cell embeddings capture Hex's neighborhood structure, is learned and converges in a similar time frame as the probing task.}
    \label{fig:ordering}
\end{figure}

Where the network body processes board configurations, MCTS directly governs the decision-making procecess. In principle, either module could be the first to discover the game concepts. For instance, the network body could start to represent concepts via updates to the value function, later enabling MCTS to successfully navigate the concept templates. We find that the first improvements in behavioral tests precede the first improvements in probing accuracy (\cref{fig:ordering}), where we consider first improvement as the epoch with a 5\% increase over the baseline value. MCTS seems to discover concepts, especially the internal concepts. Then, as the policy network is trained to match the MCTS logits, the concept representation is absorbed into the network.

We find evidence that the structure of the board is first learned at about the same time other concepts are learned. In \cref{sec:discussion:board}, we discuss how we probe for this structure, determining if AZ encodes the relative distance between all cells on the board. The last column of \cref{fig:ordering}, labeled \textit{structural}, shows that improvements on this concept occur at a similar time as to other concepts. This suggests that AZ does not learn concepts according to an obvious order or curriculum, but rather, concepts of differing levels of complexity develop in parallel.

The results shown in \cref{fig:ordering} outline the relationship between the concepts as they are learned during training, but does not establish a causal relationship between how the concept representation (as detected by the probing models) impacts how the model uses that concept.

\subsection{AlphaZero's Cell Embeddings Capture the Structure of the Board}\label{sec:discussion:board}

\noindent \begin{minipage}{0.45\textwidth}
Understanding Hex's concepts requires understanding the board's structure, that is, which cells connect to which other cells. AlphaZero (AZ), with its feed forward network architecture, does not \textit{a priori} represent this structure.
\cref{fig:positive} suggests that some information is learned before the model is able to use the concepts. A possibility is that AZ spends the initial portions of training building up a representation of the board. However, we find no evidence that the neighborhood structure is learned in the initial stages of training, rather, it appears to be learned only after the first improvements in behavioral tests (\cref{fig:ordering}). 
\end{minipage}
\hspace{0.025\textwidth}
\begin{minipage}{0.5\textwidth}
% \begin{figure}
    \centering
    \includegraphics[width=\linewidth]{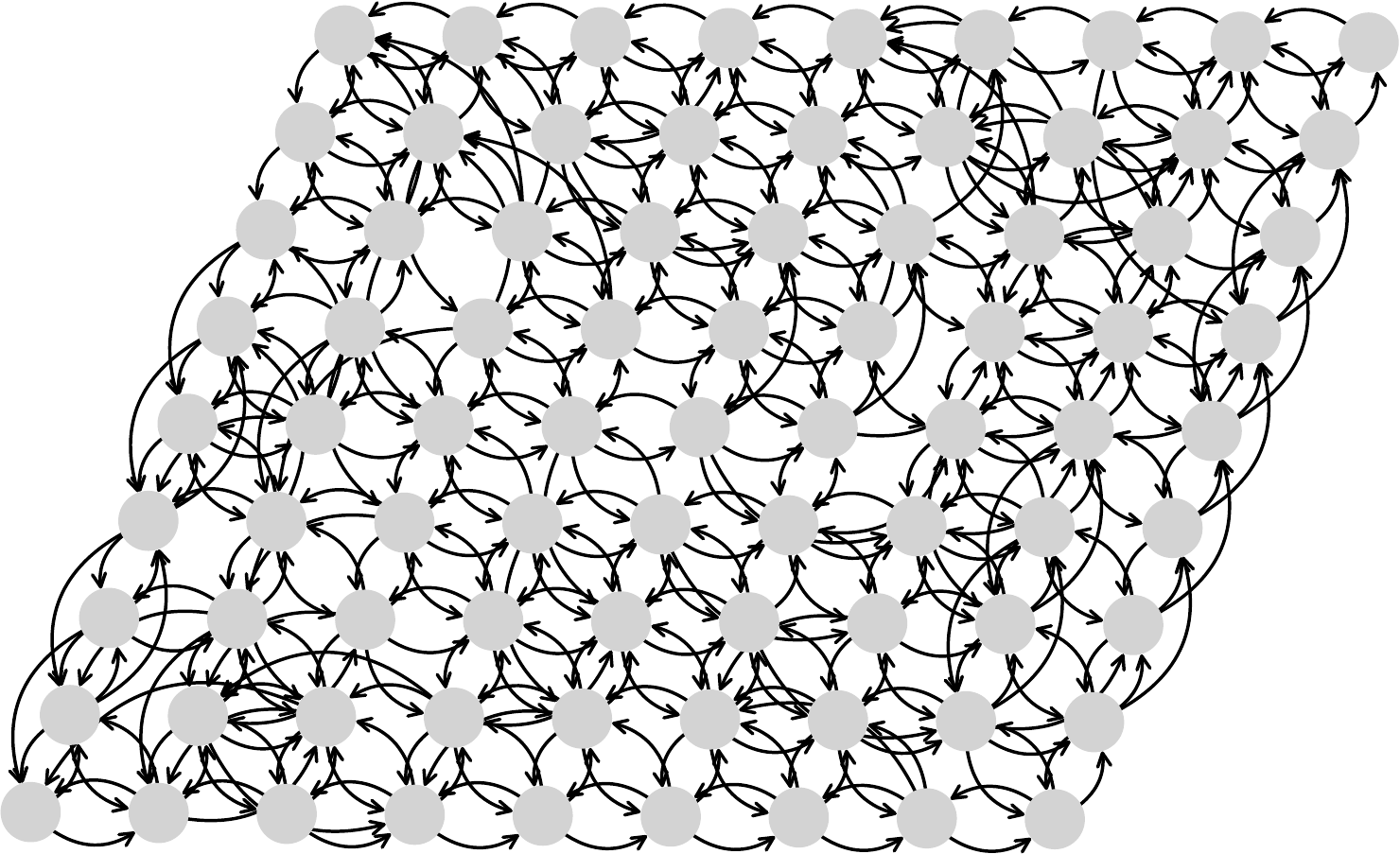}
    \captionof{figure}{\textbf{Implicit board structure.} Arrows mark the learned nearest neighbors of each cell.}
    \label{fig:board_structure}
\end{minipage}

To study if Hex's board is represented in AZ, we hypothesized that the structure of Hex's board is implicitly learned by AZ's first layer. For each Hex cell, we extract a cell embedding from the first layer of AZ.\footnote{AZ's first layer takes a flattened 1-hot encoding of the board and matrix multiplies it with a weight matrix. The embedding of each cell can be spliced out of that weight matrix. We only extract the cell embeddings for the current player. This constitutes a structural probe \cite{hewitt-manning-2019-structural}. A structural probe tests the relationship between neural network activations (or weights).} Next, we compute the dot-products between each cell embedding. The dot-product score between ground-truth neighbors increases throughout AZ's training.

The nearest-neighbors (according to the dot-product scores) nearly match the ground-truth by the 15th of 20 checkpoints, and eventually match the ground-truth before deviating slightly. We evaluate how well the dot product scores align with the ground-truth cell distances via a ranking metric, Normalized Discounted Cumulative Gain (NDCG). The NDCG first improves about 50\% of the way through training as highlighted in \cref{fig:ordering}. Details on the learned structure are in the Supplementary Material.

\section{Related Work}\label{sec:related}

\paragraph{Applying MCTS and Deep Learning to the Game of Hex.} \citet{jones2021scaling}, whose trained agents we use, studied scaling laws between various parameters, finding a mathematical relationship between compute, board size, and agent performance. This type of result can help others make informed tradeoffs when training board-game-playing agents. Where \citet{gao2017move} and \citet{takada2017reinforcement} both use value functions in concert with MCTS to play Hex, MoHex \citep{Huang2014-to} combines programmatic connection detection, pattern matching, and MCTS, attacking Hex for the 9$\times$9 grid. 

\paragraph{AlphaZero, anecdotally, uses concepts, but only achieves a partial mastery.} Domain experts have observed that AlphaZero~\citep{Silver2016-ww} and related agents that combine Monte Carlo tree search~\citep{Brugmann1993-rw} with deep reinforcement learning use identifiable concepts within board games. Michael Redmond identifies several common gameplay concepts demonstrated by AlphaZero~\citep{DeepMind2016-jk} in AlphaZero's Go matches against Lee Sedol. \citet{Silver2018-uo} noted that common human-played corner moves in Go---\textit{joseki}---were used by AlphaZero during self-play training before being abandoned, presumably as other more effective moves were found. Additional analysis provided by \citet{Tian2019-kw} note that Elf OpenGo only partially mastered ladder sequences within the game. Chess commentator Antonio Radi\'{c} \citep{Radic2017-ep} detailed how AlphaZero used \textit{zugzwang}~\citep{Winter1997-jx} in the course of defeating Stockfish~\citep{Romstad2008-dv}. Experts have already incorporated some of AlphaZero's innovations into their play~\citep{Nielsen2019-ut,sadler2019game}.

\paragraph{Probing neural networks for human concepts.} A range of linguistic concepts have been detected in NLP models using probing models \citep{conneau-etal-2018-cram,poliak2018collecting,marvin2018targeted,tenney2019you,hewitt-manning-2019-structural}. For example, \citet{tenney2019you}, with probing models, found that BERT appears to best encode linguistic concepts in the same order those concepts might be processed by the typical NLP pipeline. There has been some discussion on what good probing accuracy signifies. \citet{hewitt-liang-2019-designing} calls for baseline controls, and to measure only the gain in accuracy compared to these baselines. \citet{voita2020information} found that measures beyond accuracy, namely Minimum Description Length (MDL)---which is used to measure how easy it to detect a given concept---provided more stable results.

\paragraph{Understanding reinforcement-learning agents trained to play board games.}
\citet{sadler2019game} cover how some particulars from AZ's gameplay has impacted high-level chess. Concurrent to our work, \citet{mcgrath2021acquisition} also looked at how AZ acquires game knowledge (which we term \emph{concepts}) in chess. Using similar probing techniques, and the original (and larger) AZ, they find that it learns to encode many of the prototypical chess concepts. Related, but not focused on model understanding per se, \citet{jhamtani2018learning} collected an annotated set of chess games. This type of resource is similar to what is used by \citet{mcgrath2021acquisition} for their probing task. In our work, we focus on controlled programmatic generation of boards with (and without) concepts present. While these board instances may not occur during selfplay, the human annotated games are also off-distribution with respect to AZ. Beyond finding similar results in a different game (Hex vs chess), we also leveraged behavioral tests, tying the concept representation to agent behavior.

\section{Discussion}\label{sec:conclusion}
Our analyses suggest that AlphaZero (AZ) learned to both represent and use concepts that humans consider important when playing  Hex. However, we found that AZ is sometimes either blind or agnostic to a winning move. This had interesting ramifications. For instance, the negative concepts, especially dead cells, were not as well-encoded or used compared to other concepts. This may be because AZ has no ``qualms'' with playing wasted moves, like dead cells, if doing so doesn't change the outcome of the game.

A layerwise analysis showed that the same concept is represented most strongly in different layers, depending on its context: short-term concepts that inform actions at the end of the game are encoded in the upper layers of the model, whereas long-term concepts are absorbed deeper into the network. This absorption mirrors how AZ's policy head was trained to predict the policy outputs of MCTS. 

Combining both representational and behavioral approaches to analyze reinforcement-learning agents allows for a fuller understanding of how agents learn. Studying the representations of concepts allows us to ask (and answer) a rich set of questions about where those concepts reside. Studying the behavior of the model with respect to a given concept tests that this representation can be translated into action.  Behavioral tests can also determine whether the model may be using known heuristics. They are complementary approaches. In future work,  applying a causal approach to the study of agents' policies will further illustrate how well these agents understand these concepts. 

\section*{Acknowledgements}
We would like to thank the reviewers across different iterations of this work; they helped us clarify our work and findings. This research was conducted using computational resources and services at the Center for Computation and Visualization, Brown University. Also, special thanks to Matthew Seymour for his Hex drawing software: \url{http://www.mseymour.ca/hex_draw/hexdraw.html}.
Charles Lovering was supported in part by the DARPA GAILA program. This work received support from the ONR PERISCOPE MURI award N00014-17-1-2699.

\bibliography{forarxiv}

\begin{thebibliography}{74}
\providecommand{\natexlab}[1]{#1}
\providecommand{\url}[1]{\texttt{#1}}
\expandafter\ifx\csname urlstyle\endcsname\relax
  \providecommand{\doi}[1]{doi: #1}\else
  \providecommand{\doi}{doi: \begingroup \urlstyle{rm}\Url}\fi

\bibitem[Alain and Bengio(2017)]{Alain2017-xb}
G.~Alain and Y.~Bengio.
\newblock Understanding intermediate layers using linear classifier probes.
\newblock \emph{ICLR Workshops}, 2017.

\bibitem[Arneson et~al.(2018)Arneson, Henderson, Pawlewicz, Huang, Young, and
  Gao]{Arneson2018-oh}
B.~Arneson, P.~Henderson, J.~Pawlewicz, A.~Huang, K.~Young, and C.~Gao.
\newblock Benzene, 2018.
\newblock
  \url{https://github.com/cgao3/benzene-vanilla-cmake/blob/d450c01eb38803b1766ed9abea51568c4672f46b/src/hex/EndgameUtil.cpp}.

\bibitem[Badgeley et~al.(2019)Badgeley, Zech, Oakden-Rayner, Glicksberg, Liu,
  Gale, McConnell, Percha, Snyder, and Dudley]{Badgeley2019-yx}
M.~A. Badgeley, J.~R. Zech, L.~Oakden-Rayner, B.~S. Glicksberg, M.~Liu,
  W.~Gale, M.~V. McConnell, B.~Percha, T.~M. Snyder, and J.~T. Dudley.
\newblock Deep learning predicts hip fracture using confounding patient and
  healthcare variables.
\newblock \emph{NPJ Digit Med}, 2:\penalty0 31, Apr. 2019.

\bibitem[Belinkov(2022)]{belinkov2022probing}
Y.~Belinkov.
\newblock Probing classifiers: Promises, shortcomings, and advances.
\newblock \emph{Computational Linguistics}, 48\penalty0 (1):\penalty0 207--219,
  2022.

\bibitem[Belinkov et~al.(2020)Belinkov, Gehrmann, and
  Pavlick]{belinkov2020interpretability}
Y.~Belinkov, S.~Gehrmann, and E.~Pavlick.
\newblock Interpretability and analysis in neural nlp.
\newblock In \emph{Proceedings of the 58th annual meeting of the association
  for computational linguistics: tutorial abstracts}, pages 1--5, 2020.

\bibitem[Bellemare et~al.(2020)Bellemare, Candido, Castro, Gong, Machado,
  Moitra, Ponda, and Wang]{Bellemare2020-uu}
M.~G. Bellemare, S.~Candido, P.~S. Castro, J.~Gong, M.~C. Machado, S.~Moitra,
  S.~S. Ponda, and Z.~Wang.
\newblock Autonomous navigation of stratospheric balloons using reinforcement
  learning.
\newblock \emph{Nature}, 588\penalty0 (7836):\penalty0 77--82, Dec. 2020.

\bibitem[Bj{\"o}rnsson et~al.(2006)Bj{\"o}rnsson, Hayward, Johanson, and van
  Rijswijck]{bjornsson2006dead}
Y.~Bj{\"o}rnsson, R.~Hayward, M.~Johanson, and J.~van Rijswijck.
\newblock Dead cell analysis in {H}ex and the {S}hannon game.
\newblock In \emph{Graph Theory in Paris}, pages 45--59. Springer, 2006.

\bibitem[Br{\"u}gmann(1993)]{Brugmann1993-rw}
B.~Br{\"u}gmann.
\newblock Monte {C}arlo go.
\newblock Technical report, Max Planck Institute of Physics, 1993.

\bibitem[Conneau et~al.(2018)Conneau, Kruszewski, Lample, Barrault, and
  Baroni]{conneau-etal-2018-cram}
A.~Conneau, G.~Kruszewski, G.~Lample, L.~Barrault, and M.~Baroni.
\newblock What you can cram into a single {\$}{\&}!{\#}* vector: Probing
  sentence embeddings for linguistic properties.
\newblock In \emph{Proceedings of the 56th Annual Meeting of the Association
  for Computational Linguistics (Volume 1: Long Papers)}, pages 2126--2136,
  Melbourne, Australia, July 2018. Association for Computational Linguistics.
\newblock \doi{10.18653/v1/P18-1198}.
\newblock URL \url{https://www.aclweb.org/anthology/P18-1198}.

\bibitem[Cooper et~al.(2021)Cooper, Lu, Forde, and {others}]{Cooper2021-ul}
A.~F. Cooper, Y.~Lu, J.~Forde, and {others}.
\newblock Hyperparameter optimization is deceiving us, and how to stop it.
\newblock \emph{Adv. Neural Inf. Process. Syst.}, 2021.

\bibitem[Cooper et~al.(2022)Cooper, Moss, Laufer, and {others}]{Cooper2022-ml}
A.~F. Cooper, E.~Moss, B.~Laufer, and {others}.
\newblock Accountability in an algorithmic society: relationality,
  responsibility, and robustness in machine learning.
\newblock \emph{2022 ACM Conference on}, 2022.

\bibitem[{DeepMind}(2016)]{DeepMind2016-jk}
{DeepMind}.
\newblock Match 1---{G}oogle {DeepMind} challenge match: {L}ee {S}edol vs
  {AlphaGo}, Mar. 2016.

\bibitem[Doshi-Velez and Kim(2017)]{Doshi-Velez2017-dp}
F.~Doshi-Velez and B.~Kim.
\newblock Towards a rigorous science of interpretable machine learning.
\newblock \emph{arXiv}, Feb. 2017.

\bibitem[Ettinger(2020)]{ettinger2020bert}
A.~Ettinger.
\newblock What bert is not: Lessons from a new suite of psycholinguistic
  diagnostics for language models.
\newblock \emph{Transactions of the Association for Computational Linguistics},
  8:\penalty0 34--48, 2020.

\bibitem[Forde et~al.(2021)Forde, Cooper, Kwegyir-Aggrey, and
  {others}]{Forde2021-sk}
J.~Z. Forde, A.~F. Cooper, K.~Kwegyir-Aggrey, and {others}.
\newblock Model selection's disparate impact in {Real-World} deep learning
  applications.
\newblock \emph{arXiv preprint arXiv}, 2021.

\bibitem[Gale(1979)]{Gale1979-db}
D.~Gale.
\newblock The game of hex and the brouwer {Fixed-Point} theorem.
\newblock \emph{Am. Math. Mon.}, 86\penalty0 (10):\penalty0 818--827, Dec.
  1979.

\bibitem[Gao et~al.(2017)Gao, Hayward, and M{\"u}ller]{gao2017move}
C.~Gao, R.~Hayward, and M.~M{\"u}ller.
\newblock Move prediction using deep convolutional neural networks in hex.
\newblock \emph{IEEE Transactions on Games}, 10\penalty0 (4):\penalty0
  336--343, 2017.

\bibitem[Gardner(1961)]{Gardner1958-bp}
M.~Gardner.
\newblock \emph{The 2nd Scientific American Book of Mathematical Puzzles and
  Diversions}.
\newblock Simon \& Schuster, 1961.

\bibitem[Gauci et~al.(2018)Gauci, Conti, Liang, Virochsiri, He, Kaden,
  Narayanan, Ye, Chen, and Fujimoto]{Gauci2018-lw}
J.~Gauci, E.~Conti, Y.~Liang, K.~Virochsiri, Y.~He, Z.~Kaden, V.~Narayanan,
  X.~Ye, Z.~Chen, and S.~Fujimoto.
\newblock Horizon: Facebook's open source applied reinforcement learning
  platform.
\newblock Nov. 2018.

\bibitem[Gauthier et~al.(2020)Gauthier, Hu, Wilcox, Qian, and
  P.~Levy]{gauthier-etal:2020-syntaxgym}
J.~Gauthier, J.~Hu, E.~Wilcox, P.~Qian, and R.~P.~Levy.
\newblock Syntaxgym: An online platform for targeted evaluation of language
  models.
\newblock In \emph{Proceedings of the 58th Annual Meeting of the Association
  for Computational Linguistics}, page 70–76, 2020.

\bibitem[Geva et~al.(2020)Geva, Schuster, Berant, and
  Levy]{geva2020transformer}
M.~Geva, R.~Schuster, J.~Berant, and O.~Levy.
\newblock Transformer feed-forward layers are key-value memories.
\newblock \emph{arXiv preprint arXiv:2012.14913}, 2020.

\bibitem[Gu et~al.(2017)Gu, Holly, Lillicrap, and Levine]{Gu2017-mq}
S.~Gu, E.~Holly, T.~Lillicrap, and S.~Levine.
\newblock Deep reinforcement learning for robotic manipulation with
  asynchronous off-policy updates.
\newblock In \emph{2017 {IEEE} International Conference on Robotics and
  Automation ({ICRA})}, pages 3389--3396. ieeexplore.ieee.org, May 2017.

\bibitem[Gururangan et~al.(2018)Gururangan, Swayamdipta, Levy, Schwartz,
  Bowman, and Smith]{gururangan2018annotation}
S.~Gururangan, S.~Swayamdipta, O.~Levy, R.~Schwartz, S.~R. Bowman, and N.~A.
  Smith.
\newblock Annotation artifacts in natural language inference data.
\newblock \emph{arXiv preprint arXiv:1803.02324}, 2018.

\bibitem[Hayward et~al.(2005)Hayward, Bj{\"o}rnsson, Johanson, Kan, Po, and
  Van~Rijswijck]{hayward2005solving}
R.~Hayward, Y.~Bj{\"o}rnsson, M.~Johanson, M.~Kan, N.~Po, and J.~Van~Rijswijck.
\newblock Solving 7$\times$ 7 hex with domination, fill-in, and virtual
  connections.
\newblock \emph{Theoretical Computer Science}, 349\penalty0 (2):\penalty0
  123--139, 2005.

\bibitem[Hewitt and Liang(2019)]{hewitt-liang-2019-designing}
J.~Hewitt and P.~Liang.
\newblock Designing and interpreting probes with control tasks.
\newblock In \emph{Proceedings of the 2019 Conference on Empirical Methods in
  Natural Language Processing and the 9th International Joint Conference on
  Natural Language Processing (EMNLP-IJCNLP)}, pages 2733--2743, Hong Kong,
  China, Nov. 2019. Association for Computational Linguistics.
\newblock \doi{10.18653/v1/D19-1275}.
\newblock URL \url{https://www.aclweb.org/anthology/D19-1275}.

\bibitem[Hewitt and Manning(2019)]{hewitt-manning-2019-structural}
J.~Hewitt and C.~D. Manning.
\newblock {A} structural probe for finding syntax in word representations.
\newblock In \emph{Proceedings of the 2019 Conference of the North {A}merican
  Chapter of the Association for Computational Linguistics: Human Language
  Technologies, Volume 1 (Long and Short Papers)}, pages 4129--4138,
  Minneapolis, Minnesota, June 2019. Association for Computational Linguistics.
\newblock \doi{10.18653/v1/N19-1419}.
\newblock URL \url{https://www.aclweb.org/anthology/N19-1419}.

\bibitem[Hu et~al.()Hu, Gauthier, Qian, Wilcox, and
  Levy]{hu-etal-2020-systematic}
J.~Hu, J.~Gauthier, P.~Qian, E.~Wilcox, and R.~Levy.
\newblock A systematic assessment of syntactic generalization in neural
  language models.
\newblock In \emph{Proc{.~}of NAACL}.
\newblock URL \url{https://www.aclweb.org/anthology/2020.acl-main.158}.

\bibitem[Huang et~al.(2014)Huang, Arneson, Hayward, M{\"u}ller, and
  Pawlewicz]{Huang2014-to}
S.-C. Huang, B.~Arneson, R.~B. Hayward, M.~M{\"u}ller, and J.~Pawlewicz.
\newblock {MoHex} 2.0: A {Pattern-Based} {MCTS} hex player.
\newblock In \emph{Computers and Games}, pages 60--71. Springer International
  Publishing, 2014.

\bibitem[Jhamtani et~al.(2018)Jhamtani, Gangal, Hovy, Neubig, and
  Berg-Kirkpatrick]{jhamtani2018learning}
H.~Jhamtani, V.~Gangal, E.~Hovy, G.~Neubig, and T.~Berg-Kirkpatrick.
\newblock Learning to generate move-by-move commentary for chess games from
  large-scale social forum data.
\newblock In \emph{Proceedings of the 56th Annual Meeting of the Association
  for Computational Linguistics (Volume 1: Long Papers)}, pages 1661--1671,
  2018.

\bibitem[Jones(2021{\natexlab{a}})]{Jones2021-sb}
A.~Jones.
\newblock {LICENSE} at master · andyljones/boardlaw, 2021{\natexlab{a}}.

\bibitem[Jones(2021{\natexlab{b}})]{jones2021scaling}
A.~L. Jones.
\newblock Scaling scaling laws with board games.
\newblock \emph{arXiv preprint arXiv:2104.03113}, 2021{\natexlab{b}}.

\bibitem[Jumper et~al.(2021)Jumper, Evans, Pritzel, Green, Figurnov,
  Ronneberger, Tunyasuvunakool, Bates, {\v{Z}}{\'\i}dek, Potapenko,
  et~al.]{jumper2021highly}
J.~Jumper, R.~Evans, A.~Pritzel, T.~Green, M.~Figurnov, O.~Ronneberger,
  K.~Tunyasuvunakool, R.~Bates, A.~{\v{Z}}{\'\i}dek, A.~Potapenko, et~al.
\newblock Highly accurate protein structure prediction with alphafold.
\newblock \emph{Nature}, 596\penalty0 (7873):\penalty0 583--589, 2021.

\bibitem[Kim and Linzen(2019)]{kim2019compositionality}
N.~Kim and T.~Linzen.
\newblock Compositionality as directional consistency in sequential neural
  networks.
\newblock In \emph{Workshop on Context and Compositionality in Biological and
  Artificial Neural Systems}, 2019.

\bibitem[King(2004)]{King2004-fg}
D.~King.
\newblock Hall of hexagons.
\newblock \url{https://www.drking.org.uk/hexagons/index.html}, 2004.
\newblock Accessed: 2021-11-12.

\bibitem[Lake and Baroni(2018)]{lake2018generalization}
B.~Lake and M.~Baroni.
\newblock Generalization without systematicity: On the compositional skills of
  sequence-to-sequence recurrent networks.
\newblock In \emph{Proc{.~}of ICML}, 2018.
\newblock URL \url{https://arxiv.org/abs/1711.00350}.

\bibitem[Linzen et~al.(2016{\natexlab{a}})Linzen, Dupoux, and
  Goldberg]{linzen-etal-2016-assessing}
T.~Linzen, E.~Dupoux, and Y.~Goldberg.
\newblock Assessing the ability of {LSTM}s to learn syntax-sensitive
  dependencies.
\newblock \emph{Transactions of the Association for Computational Linguistics},
  4:\penalty0 521--535, 2016{\natexlab{a}}.
\newblock \doi{10.1162/tacl_a_00115}.
\newblock URL \url{https://www.aclweb.org/anthology/Q16-1037}.

\bibitem[Linzen et~al.(2016{\natexlab{b}})Linzen, Dupoux, and
  Goldberg]{linzen2016assessing}
T.~Linzen, E.~Dupoux, and Y.~Goldberg.
\newblock {Assessing the Ability of LSTMs to Learn Syntax-Sensitive
  Dependencies}.
\newblock \emph{Transactions of the Association for Computational Linguistics},
  4:\penalty0 521--535, 2016{\natexlab{b}}.
\newblock URL \url{http://aclweb.org/anthology/Q16-1037}.

\bibitem[Lovering et~al.(2021)Lovering, Jha, Linzen, and
  Pavlick]{lovering2021predicting}
C.~Lovering, R.~Jha, T.~Linzen, and E.~Pavlick.
\newblock Predicting inductive biases of pre-trained models, 2021.

\bibitem[Lucic et~al.(2022)Lucic, Bleeker, Bhargav, Forde, Sinha, Dodge,
  Luccioni, and Stojnic]{Lucic2022-fa}
A.~Lucic, M.~Bleeker, S.~Bhargav, J.~Forde, K.~Sinha, J.~Dodge, S.~Luccioni,
  and R.~Stojnic.
\newblock Towards reproducible machine learning research in natural language
  processing.
\newblock In \emph{Proceedings of the 60th Annual Meeting of the Association
  for Computational Linguistics: Tutorial Abstracts}, pages 7--11, Dublin,
  Ireland, May 2022. Association for Computational Linguistics.

\bibitem[Margolis et~al.(1999)Margolis, Laurence, et~al.]{margolis1999concepts}
E.~Margolis, S.~Laurence, et~al.
\newblock \emph{Concepts: core readings}.
\newblock Mit Press, 1999.

\bibitem[Marvin and Linzen(2018)]{marvin2018targeted}
R.~Marvin and T.~Linzen.
\newblock Targeted syntactic evaluation of language models.
\newblock \emph{arXiv preprint arXiv:1808.09031}, 2018.

\bibitem[McCoy et~al.(2020)McCoy, Pavlick, and Linzen]{mccoy2020right}
R.~T. McCoy, E.~Pavlick, and T.~Linzen.
\newblock Right for the wrong reasons: Diagnosing syntactic heuristics in
  natural language inference.
\newblock In \emph{57th Annual Meeting of the Association for Computational
  Linguistics, ACL 2019}, pages 3428--3448. Association for Computational
  Linguistics (ACL), 2020.

\bibitem[McGrath et~al.(2021)McGrath, Kapishnikov, Toma{\v{s}}ev, Pearce,
  Hassabis, Kim, Paquet, and Kramnik]{mcgrath2021acquisition}
T.~McGrath, A.~Kapishnikov, N.~Toma{\v{s}}ev, A.~Pearce, D.~Hassabis, B.~Kim,
  U.~Paquet, and V.~Kramnik.
\newblock Acquisition of chess knowledge in alphazero.
\newblock \emph{arXiv preprint arXiv:2111.09259}, 2021.

\bibitem[Meng et~al.(2022)Meng, Bau, Andonian, and Belinkov]{meng2022locating}
K.~Meng, D.~Bau, A.~Andonian, and Y.~Belinkov.
\newblock Locating and editing factual knowledge in gpt.
\newblock \emph{arXiv preprint arXiv:2202.05262}, 2022.

\bibitem[Mirhoseini et~al.(2021)Mirhoseini, Goldie, Yazgan, Jiang, Songhori,
  Wang, Lee, Johnson, Pathak, Nazi, Pak, Tong, Srinivasa, Hang, Tuncer, Le,
  Laudon, Ho, Carpenter, and Dean]{Mirhoseini2021-ni}
A.~Mirhoseini, A.~Goldie, M.~Yazgan, J.~W. Jiang, E.~Songhori, S.~Wang, Y.-J.
  Lee, E.~Johnson, O.~Pathak, A.~Nazi, J.~Pak, A.~Tong, K.~Srinivasa, W.~Hang,
  E.~Tuncer, Q.~V. Le, J.~Laudon, R.~Ho, R.~Carpenter, and J.~Dean.
\newblock A graph placement methodology for fast chip design.
\newblock \emph{Nature}, 594\penalty0 (7862):\penalty0 207--212, June 2021.

\bibitem[Nielsen(2019)]{Nielsen2019-ut}
P.~H. Nielsen.
\newblock The exciting impact of a game changer: When {M}agnus met {AlphaZero}.
\newblock \emph{New In Chess}, 2019.

\bibitem[Ormerod(2016)]{omerod2016}
D.~Ormerod.
\newblock \emph{AlphaGo shows its true strength in 3rd victory against Lee
  Sedol}.
\newblock 2016.

\bibitem[Pandia and Ettinger(2021)]{pandia2021sorting}
L.~Pandia and A.~Ettinger.
\newblock Sorting through the noise: Testing robustness of information
  processing in pre-trained language models.
\newblock In \emph{EMNLP (1)}, 2021.

\bibitem[Pawlewicz and Hayward(2014)]{Pawlewicz2014-ca}
J.~Pawlewicz and R.~B. Hayward.
\newblock Scalable parallel {DFPN} search.
\newblock In \emph{Computers and Games}, pages 138--150. Springer International
  Publishing, 2014.

\bibitem[Pawlewicz et~al.(2014)Pawlewicz, Hayward, Henderson, and
  Arneson]{pawlewicz2014stronger}
J.~Pawlewicz, R.~Hayward, P.~Henderson, and B.~Arneson.
\newblock Stronger virtual connections in hex.
\newblock \emph{IEEE Transactions on Computational Intelligence and AI in
  Games}, 7\penalty0 (2):\penalty0 156--166, 2014.

\bibitem[Pawlewicz et~al.(2015)Pawlewicz, Hayward, Henderson, and
  Arneson]{Pawlewicz2015-vc}
J.~Pawlewicz, R.~Hayward, P.~Henderson, and B.~Arneson.
\newblock Stronger virtual connections in hex.
\newblock \emph{IEEE Trans. Comput. Intell. AI Games}, 7\penalty0 (2):\penalty0
  156--166, June 2015.

\bibitem[Poliak et~al.(2018{\natexlab{a}})Poliak, Haldar, Rudinger, Hu,
  Pavlick, White, and Van~Durme]{poliak2018collecting}
A.~Poliak, A.~Haldar, R.~Rudinger, J.~E. Hu, E.~Pavlick, A.~S. White, and
  B.~Van~Durme.
\newblock Collecting diverse natural language inference problems for sentence
  representation evaluation.
\newblock \emph{arXiv preprint arXiv:1804.08207}, 2018{\natexlab{a}}.

\bibitem[Poliak et~al.(2018{\natexlab{b}})Poliak, Naradowsky, Haldar, Rudinger,
  and Van~Durme]{poliak2018hypothesis}
A.~Poliak, J.~Naradowsky, A.~Haldar, R.~Rudinger, and B.~Van~Durme.
\newblock Hypothesis only baselines in natural language inference.
\newblock \emph{arXiv preprint arXiv:1805.01042}, 2018{\natexlab{b}}.

\bibitem[Popova et~al.(2018)Popova, Isayev, and Tropsha]{Popova2018-jg}
M.~Popova, O.~Isayev, and A.~Tropsha.
\newblock Deep reinforcement learning for de novo drug design.
\newblock \emph{Science Advances}, 4\penalty0 (7):\penalty0 eaap7885, July
  2018.

\bibitem[Radi{\'c}(2017)]{Radic2017-ep}
A.~Radi{\'c}.
\newblock {AlphaZero}'s ``immortal zugzwang game'' against {S}tockfish, 2017.
\newblock URL \url{https://www.youtube.com/watch?v=lFXJWPhDsSY}.

\bibitem[Romstad et~al.(2008)Romstad, Costalba, Kiiski, and
  {others}]{Romstad2008-dv}
T.~Romstad, M.~Costalba, J.~Kiiski, and {others}.
\newblock Stockfish, 2008.

\bibitem[Sadler and Regan(2019)]{sadler2019game}
M.~Sadler and N.~Regan.
\newblock Game changer.
\newblock \emph{AlphaZero’s Groundbreaking Chess Strategies and the Promise
  of AI. Alkmaar. The Netherlands. New in Chess}, 2019.

\bibitem[Seymour(2019)]{Seymour2019-hy}
M.~Seymour.
\newblock \emph{Hex: A Strategy Guide}.
\newblock 2019.
\newblock URL \url{http://www.mseymour.ca/hex_book/hexstrat.html}.

\bibitem[Silver et~al.(2016)Silver, Huang, Maddison, Guez, Sifre, van~den
  Driessche, Schrittwieser, Antonoglou, Panneershelvam, Lanctot, Dieleman,
  Grewe, Nham, Kalchbrenner, Sutskever, Lillicrap, Leach, Kavukcuoglu, Graepel,
  and Hassabis]{Silver2016-ww}
D.~Silver, A.~Huang, C.~J. Maddison, A.~Guez, L.~Sifre, G.~van~den Driessche,
  J.~Schrittwieser, I.~Antonoglou, V.~Panneershelvam, M.~Lanctot, S.~Dieleman,
  D.~Grewe, J.~Nham, N.~Kalchbrenner, I.~Sutskever, T.~Lillicrap, M.~Leach,
  K.~Kavukcuoglu, T.~Graepel, and D.~Hassabis.
\newblock Mastering the game of go with deep neural networks and tree search.
\newblock \emph{Nature}, 529\penalty0 (7587):\penalty0 484--489, Jan. 2016.

\bibitem[Silver et~al.(2018)Silver, Hubert, Schrittwieser, Antonoglou, Lai,
  Guez, Lanctot, Sifre, Kumaran, Graepel, Lillicrap, Simonyan, and
  Hassabis]{Silver2018-uo}
D.~Silver, T.~Hubert, J.~Schrittwieser, I.~Antonoglou, M.~Lai, A.~Guez,
  M.~Lanctot, L.~Sifre, D.~Kumaran, T.~Graepel, T.~Lillicrap, K.~Simonyan, and
  D.~Hassabis.
\newblock A general reinforcement learning algorithm that masters chess, shogi,
  and go through self-play.
\newblock \emph{Science}, 362\penalty0 (6419):\penalty0 1140--1144, Dec. 2018.

\bibitem[Takada et~al.(2017)Takada, Iizuka, and
  Yamamoto]{takada2017reinforcement}
K.~Takada, H.~Iizuka, and M.~Yamamoto.
\newblock Reinforcement learning for creating evaluation function using
  convolutional neural network in hex.
\newblock In \emph{2017 Conference on Technologies and Applications of
  Artificial Intelligence (TAAI)}, pages 196--201. IEEE, 2017.

\bibitem[Tenney et~al.(2019{\natexlab{a}})Tenney, Das, and
  Pavlick]{tenney-etal-2019-bert}
I.~Tenney, D.~Das, and E.~Pavlick.
\newblock {BERT} rediscovers the classical {NLP} pipeline.
\newblock In \emph{Proceedings of the 57th Annual Meeting of the Association
  for Computational Linguistics}, pages 4593--4601, Florence, Italy, July
  2019{\natexlab{a}}. Association for Computational Linguistics.
\newblock \doi{10.18653/v1/P19-1452}.
\newblock URL \url{https://aclanthology.org/P19-1452}.

\bibitem[Tenney et~al.(2019{\natexlab{b}})Tenney, Xia, Chen, Wang, Poliak,
  McCoy, Kim, Durme, Bowman, Das, and Pavlick]{tenney2019you}
I.~Tenney, P.~Xia, B.~Chen, A.~Wang, A.~Poliak, R.~T. McCoy, N.~Kim, B.~V.
  Durme, S.~Bowman, D.~Das, and E.~Pavlick.
\newblock {What do you learn from context? Probing for sentence structure in
  contextualized word representations}.
\newblock In \emph{International Conference on Learning Representations},
  2019{\natexlab{b}}.
\newblock URL \url{https://openreview.net/forum?id=SJzSgnRcKX}.

\bibitem[Tian et~al.(2019)Tian, Ma, Gong, Sengupta, Chen, Pinkerton, and
  Zitnick]{Tian2019-kw}
Y.~Tian, J.~Ma, Q.~Gong, S.~Sengupta, Z.~Chen, J.~Pinkerton, and L.~Zitnick.
\newblock Elf opengo: An analysis and open reimplementation of alphazero.
\newblock In \emph{International Conference on Machine Learning}, pages
  6244--6253, 2019.

\bibitem[Tomlin et~al.(2022)Tomlin, He, and Klein]{Tomlin2022}
N.~Tomlin, A.~He, and D.~Klein.
\newblock Understanding game-playing agents with natural language annotations.
\newblock In \emph{Proceedings of the 60th Annual Meeting of the Association
  for Computational Linguistics (Volume 2: Short Papers)}, pages 797--807,
  2022.

\bibitem[Voita and Titov(2020)]{voita2020information}
E.~Voita and I.~Titov.
\newblock Information-theoretic probing with minimum description length.
\newblock In \emph{Proceedings of the 2020 Conference on Empirical Methods in
  Natural Language Processing (EMNLP)}, pages 183--196, Online, Nov. 2020.
  Association for Computational Linguistics.
\newblock \doi{10.18653/v1/2020.emnlp-main.14}.
\newblock URL \url{https://www.aclweb.org/anthology/2020.emnlp-main.14}.

\bibitem[Warstadt et~al.(2020{\natexlab{a}})Warstadt, Parrish, Liu, Mohananey,
  Peng, Wang, and Bowman]{warstadt2020blimp}
A.~Warstadt, A.~Parrish, H.~Liu, A.~Mohananey, W.~Peng, S.-F. Wang, and S.~R.
  Bowman.
\newblock {BLiMP: The benchmark of linguistic minimal pairs for English}.
\newblock \emph{Transactions of the Association for Computational Linguistics},
  8:\penalty0 377--392, 2020{\natexlab{a}}.

\bibitem[Warstadt et~al.(2020{\natexlab{b}})Warstadt, Zhang, Li, Liu, and
  Bowman]{warstadt-etal-2020-learning}
A.~Warstadt, Y.~Zhang, X.~Li, H.~Liu, and S.~R. Bowman.
\newblock Learning which features matter: {R}o{BERT}a acquires a preference for
  linguistic generalizations (eventually).
\newblock In \emph{Proc{.~}of EMNLP}, 2020{\natexlab{b}}.
\newblock URL \url{https://www.aclweb.org/anthology/2020.emnlp-main.16}.

\bibitem[Winter(1997)]{Winter1997-jx}
E.~Winter.
\newblock Zugzwang.
\newblock \url{https://www.chesshistory.com/winter/extra/zugzwang.html}, 1997.
\newblock Accessed: 2021-11-12.

\bibitem[Witty et~al.(2018)Witty, Lee, Tosch, Atrey, Littman, and
  Jensen]{Witty2018-sb}
S.~Witty, J.~K. Lee, E.~Tosch, A.~Atrey, M.~Littman, and D.~Jensen.
\newblock Measuring and characterizing generalization in deep reinforcement
  learning.
\newblock \emph{arXiv preprint arXiv:1812.02868}, 2018.

\bibitem[Witty et~al.(2021)Witty, Lee, Tosch, Atrey, Clary, Littman, and
  Jensen]{Witty2021-ge}
S.~Witty, J.~K. Lee, E.~Tosch, A.~Atrey, K.~Clary, M.~L. Littman, and
  D.~Jensen.
\newblock Measuring and characterizing generalization in deep reinforcement
  learning.
\newblock \emph{Applied AI Letters}, 2\penalty0 (4):\penalty0 e45, Dec. 2021.

\bibitem[Zech et~al.(2019{\natexlab{a}})Zech, Forde, Titano, Kaji, Costa, and
  Oermann]{Zech2019-oa}
J.~Zech, J.~Forde, J.~J. Titano, D.~Kaji, A.~Costa, and E.~K. Oermann.
\newblock Detecting insertion, substitution, and deletion errors in radiology
  reports using neural sequence-to-sequence models.
\newblock \emph{Ann Transl Med}, 7\penalty0 (11):\penalty0 233, June
  2019{\natexlab{a}}.

\bibitem[Zech et~al.(2019{\natexlab{b}})Zech, Forde, and Littman]{Zech2019-wj}
J.~R. Zech, J.~Z. Forde, and M.~L. Littman.
\newblock Individual predictions matter: Assessing the effect of data ordering
  in training fine-tuned {CNNs} for medical imaging.
\newblock Dec. 2019{\natexlab{b}}.

\bibitem[Zhang et~al.(2018)Zhang, Wu, and Pineau]{Zhang2018-ug}
A.~Zhang, Y.~Wu, and J.~Pineau.
\newblock Natural environment benchmarks for reinforcement learning.
\newblock \emph{arXiv preprint arXiv:1811.06032}, 2018.

\end{thebibliography}
\bibliographystyle{abbrvnat}

%%%%%%%%%%%%%%%%%%%%%%%%%%%%%%%%%%%%%%%%%%%%%%%%%%%%%%%%%%%%

\section*{Checklist}

\begin{enumerate}

\item For all authors...
\begin{enumerate}
  \item Do the main claims made in the abstract and introduction accurately reflect the paper's contributions and scope?
    \answerYes{}
  \item Did you describe the limitations of your work?
    \answerYes{} See \cref{sub:limitations}
  \item Did you discuss any potential negative societal impacts of your work?
    \answerYes{} See \cref{sub:impact}
  \item Have you read the ethics review guidelines and ensured that your paper conforms to them?
    \answerYes{} 
\end{enumerate}

\item If you are including theoretical results...
\begin{enumerate}
  \item Did you state the full set of assumptions of all theoretical results?
    \answerNA{}
        \item Did you include complete proofs of all theoretical results?
    \answerNA{}
\end{enumerate}

\item If you ran experiments...
\begin{enumerate}
  \item Did you include the code, data, and instructions needed to reproduce the main experimental results (either in the supplemental material or as a URL)?
    \answerYes{} See \ref{sub:reproducibility}
  \item Did you specify all the training details (e.g., data splits, hyperparameters, how they were chosen)?
    \answerYes{} See Section \ref{sec:representation}
        \item Did you report error bars (e.g., with respect to the random seed after running experiments multiple times)?
    \answerYes{} We report one standard deviation above and below the mean in \cref{fig:selectivity} and \cref{fig:ordering}.
        \item Did you include the total amount of compute and the type of resources used (e.g., type of GPUs, internal cluster, or cloud provider)?
    \answerYes{} Our hardware used and compute estimates is described in \cref{sub:reproducibility}. We will include a breakdown of compute costs in the Supplementary Materials.
\end{enumerate}

\item If you are using existing assets (e.g., code, data, models) or curating/releasing new assets...
\begin{enumerate}
  \item If your work uses existing assets, did you cite the creators?
    \answerYes{} We acknowledge the work of \citet{jones2021scaling} in \cref{sec:designing}
  \item Did you mention the license of the assets?
    \answerYes{} See \cref{sec:designing}.
  \item Did you include any new assets either in the supplemental material or as a URL?
    \answerYes{} See \cref{sub:reproducibility}
  \item Did you discuss whether and how consent was obtained from people whose data you're using/curating?
    \answerNA{}
  \item Did you discuss whether the data you are using/curating contains personally identifiable information or offensive content?
    \answerYes{} See \cref{sub:impact}
\end{enumerate}

\item If you used crowdsourcing or conducted research with human subjects...
\begin{enumerate}
  \item Did you include the full text of instructions given to participants and screenshots, if applicable?
    \answerNA{}
  \item Did you describe any potential participant risks, with links to Institutional Review Board (IRB) approvals, if applicable?
    \answerNA{}
  \item Did you include the estimated hourly wage paid to participants and the total amount spent on participant compensation?
    \answerNA{}
\end{enumerate}

\end{enumerate}

\newpage
\clearpage
\appendix
\onecolumn

\section*{Supplementary Materials}

\cref{app:sec:impl} reports implementation details, hyperparameters and compute requirements.\\
\cref{app:sec:concepts} gives more details on each concept introduced in the main body of the paper.\\
\cref{app:sec:gameplay} demonstrates how AlphaZero often wastes moves. \\
\cref{app:sec:results} has additional results across the different architectures.

\section{Implementation Details}\label{app:sec:impl}
We use agents trained by \citet{jones2021scaling}. See \cref{app:tbl:agents} for hyperparameters and relative agent strengths. We used  NVIDIA GeForce RTX 3090, to generate boards and encode them. The compute is reported in \cref{app:tbl:time} For all results in the main body of the paper we use the model \texttt{grubby}. We report additional results with other models below in \cref{app:sec:results}. 
The code, results and examples can be found at \url{https://bit.ly/alphatology}.

\begin{table*}[h]
\centering
\small
\begin{tabular}{lp{1.5cm}ccccp{1cm}c}
\toprule
 agent run & \citet{jones2021scaling} full agent run name  & depth & width & MCTS nodes & train ckpts & win rate against MoHex as black & Elo\\
%  agent run & Jones et al code  & depth & width & MCTS Nodes &  trn ckpt & win rate against MoHex as black & Elo\\
\midrule
\textbf{\texttt{grubby}} &\textbf{2021-02-20 22-35-41 grubby-wrench}    &     \textbf{8 layers}     &   \textbf{512 neurons} & \textbf{64} & \textbf{20} & \textbf{0.922} &        \textbf{-0.345} \\
\midrule
{\texttt{recent}} &  2021-02-20 21-33-42 recent-annex   &   8 layers        &   256 neurons & 64 & 19 & 0.77 & -0.361    \\
{\texttt{baggy}} &  2021-02-20 22-18-43 baggy-cans       &     4 layers     &    512 neurons & 64 & 20 & 0.922  &     -0.388  \\
{\texttt{vital}} & 2021-02-20 22-55-43 vital-bubble        & 2 layers           &   1024 neurons &  64 & 20 & 0.922 &  -0.400      \\
\bottomrule
\end{tabular}
\caption{\textbf{Agent hyperparameters.} The win rate is the agent's win rate \textit{as black} vs MoHex \citep{Huang2014-to} \textit{without the swap rule}. Under perfect play in Hex, black cannot lose. The Elo rating for each agent is calculated based on trials against MoHex and against the subset of agent runs \citet{jones2021scaling} evaluated against against MoHex. Because \citet{jones2021scaling} fixes the number of MCTS nodes used to compare against MoHex at 64, we do not consider alternate agent configurations that expand or contract the number of nodes when calculating the Elo rating.}
\label{app:tbl:agents}
\end{table*}

\begin{table*}[h]
\centering
\small
\begin{tabular}{lcccc|c}
\toprule
script & time & concepts & seeds & parallel & total \\
\midrule
\texttt{encode} & 8.34 $\pm$ 4 m & x9 & x3 & - & 216 m \\
\texttt{probing} & 21.75 $\pm$ 16.35 m & x9 & x3 & /4 & 146 m \\
\texttt{positive} & 14.31 $\pm$ 8.23 m & x7 & x3 & /4 & 75 m \\
\texttt{negative} & 23.48 $\pm$ 5.46 m & x2 & x3 & /4 & 35 m \\
\bottomrule
\end{tabular}
\caption{\textbf{Compute.} For \texttt{grubby} model, the total is about 8GPU hours. The other models take less time than this model. The total run time is about 24GPU hours.  We used  NVIDIA GeForce RTX 3090.  }
\label{app:tbl:time}
\end{table*}

\begin{table*}[h]
\centering
\begin{tabular}{llrrrr}
\toprule
  experiment & parameter & $N$\\
\midrule
probing & training examples & O(2000)\\
     & test examples & O(500)\\
     & seeds & 3\\
behavioral tests & examples & 100\\
     & seeds & 3\\
\bottomrule
\end{tabular}
\caption{\textbf{Experiment hyperparameters} per concept.}
\label{app:tbl:hyper}
\end{table*}

\clearpage

\begin{figure*}[h!]
    \centering
    \includegraphics[width=\linewidth]{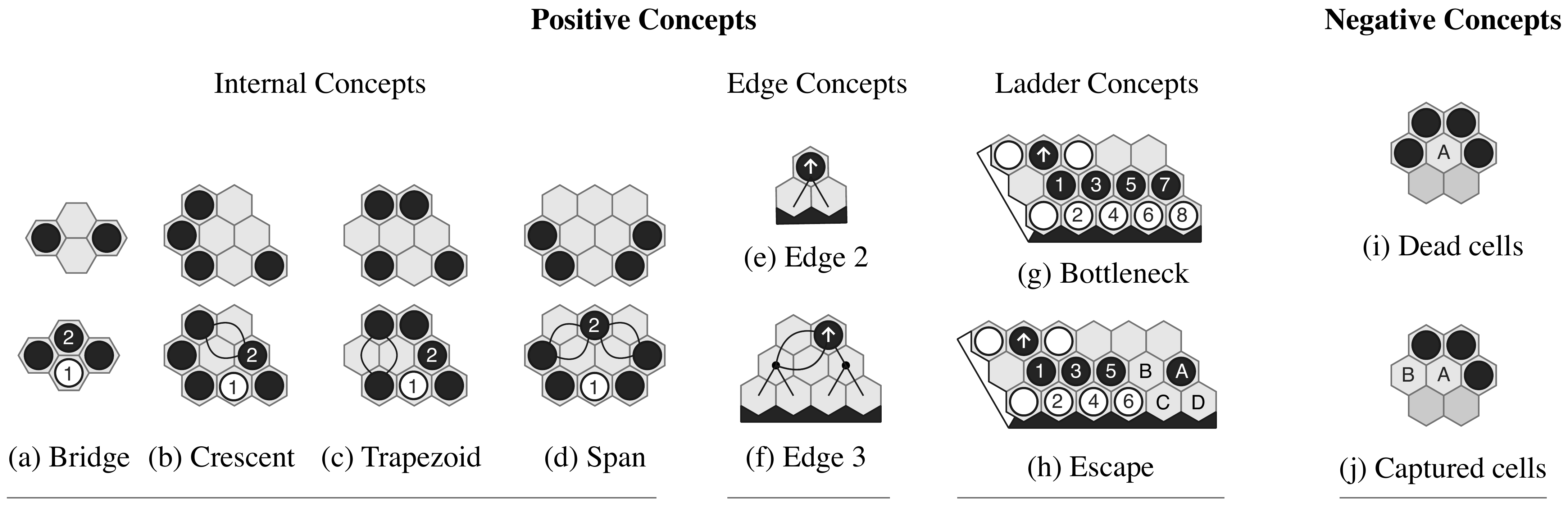} 
\caption{
\textbf{Hex templates exemplifying game concepts.} Concepts within the game of Hex are templates on the board formed by a player's pieces with known strategic and tactical implications. Positive concepts provide the player with the concept with multiple ways to connect the pieces within the concept together, despite possible attacks from the opponent.  An example of these properties is the bridge (a). If white plays move 1, black can connect the two pieces of the bridge by playing move 2. Negative concepts change the strategic value of specific open spots of the board, such that the opponent is disincentived to play those open spots, such as move A in (i). Each concept is further described in our Supplementary Material. Arrows indicate that the piece is connected to the opposite side of the board; the lines show the bridge concept within the other concepts.
}         \label{app:fig:concepts}
     \end{figure*}
     
\section{Concepts in Hex}\label{app:sec:concepts}
\paragraph{Internal Concepts}
With the goal of Hex being to build a chain across the board, it is helpful to recognize when cells are virtually connected, that is, even in response to perfect adversarial play, the cells are guaranteed to connect \citep{hayward2005solving,pawlewicz2014stronger}. The bridge (\cref{app:fig:concepts}a) is the simplest such concept. The larger internal templates -- crescent, trapezoid, span  (\cref{app:fig:concepts}(b,c,d)) -- provide several possibilities to connect a player's pieces. 

\paragraph{Edge Concepts} 
An edge template guarantees a connection from a single cell to a given edge \citep{King2004-fg,Seymour2019-hy}. Recognizing these edge templates is important for building effective strategies. In this work, we consider the edge concept to be the two templates shown in \cref{app:fig:concepts}(e, f). In \cref{app:fig:concepts}(e, f), black can connect to the bottom wall irrespective of how white attempts to sever the connection. 

\paragraph{Ladder Concepts}
Ladders are common in Hex. While similar in spirit to ladders in Go, there are some technical differences. We analyze two different ladder concepts. \textit{Bottlenecks} are a defensive concept that leads to the ladder. The defender holds off the attacker (in \cref{app:fig:concepts}g white successfully defends against black). If the attacking player (here black) continues the ladder, the defending player (here white) must block each move or lose the game. However, the defending player will eventually win, as each defensive move builds up a chain across the board. Ladder \textit{escapes} have a different outcome. If there is already a cell in path of the ladder, like A in \cref{app:fig:concepts}h, then the attacking player will win. When black plays B connecting to A, white cannot defend against both C and D.

\paragraph{Dead Cells (Useless Triangles).}
Some empty cells cannot impact the outcome of the game regardless of with which color they are filled. A cell being dead is a global fact of the board, and is difficult in general to compute \citep{bjornsson2006dead}. However, there are known templates where it is relatively easy to deduce that a cell is dead, and it is these templates with which we test the model. Among the simplest is the ``useless triangle'', which is the precursor of the more general notion of dead cells (\cref{app:fig:concepts}e). If white plays A, she does not restrict black's territory. To do so she would have to play both other empty cells, after which her move into A would not have any use. For black, playing A doesn't hinder white, nor provide any new territory.

\paragraph{Captured Cells.}
An empty cell is considered captured when it is effectively filled by a player. The templates have (at least) two empty cells A and B  (\cref{app:fig:concepts}f). If the cells are black-captured, it means that if white intrudes into the template (playing A), then black can respond by playing B and making A dead. % So, captured cells are derivative of dead cells.

\clearpage

\section{Characterizing AlphaZero's Gameplay of Hex}\label{app:sec:gameplay}

AlphaZero (AZ) uses both deep learning and MCTS.  The deep network, $\textit{f}_{\theta}({s})$, takes in boardgame input, $s$, and produces two forms of output: the value estimate of the board, $v$, and a move prior, $\textbf{p}$. These prior probabilities are then used for MCTS, which outputs the MCTS probability distribution over actions, $\boldsymbol{\pi}$. In the game of Hex, rewards, $z$, and value estimates are scaled within $[-1, 1]$ and a reward of $-1$ or $1$ is only recorded to the loser or winner of a match \citep{jones2021scaling}. AZ is optimized to minimize the mean square error of the value estimate and reward,  $v$ and $z$, and the cross entropy loss of the policy outputs of the deep network and MCTS, $\textbf{p}$ and $\boldsymbol{\pi}$:

$$l = (z-v)^2 - \boldsymbol{\pi}^{T}\log{\textbf{p}} + c||\theta||^2$$

MCTS balances exploration, search, and value estimates to pick the action that leads to the highest probability of winning. The loss does not intrude any term (or discount factor) to encourage winning more quickly. Consequently, the value estimates and action probabilities produced by the deep network and MCTS solely consider the actions that will eventually result in a win, \textit{regardless of the number of moves necessary to achieve that win}. This is in contrast to similar Hex-playing agents such as MoHex \citep{Huang2014-to, Pawlewicz2014-ca, Pawlewicz2015-vc}, which is hard-coded to connect the winning player's pieces \citep{Arneson2018-oh}.

MoHex has been demonstrated to play perfectly on boards up to 9x9, and the AZ agent we utilize has been demonstrated to play competitively with MoHex \citep{jones2021scaling}.  We observe, however, that AZ does not always end the game in the fewest possible moves.

In practice, AZ sometimes delays winning. Figure \ref{app:fig:long_endgame1} provides a hand-derived example of AZ delaying the end of a game. Black has nearly won the game, and must play its next move.  By playing B, it can end the game; because B forms the shortest connection for black, MoHex is hard-coded to select B. When AZ's MCTS is run 100 times on this board, however, AZ places higher probability on selecting A a majority of those games (Figure \ref{app:fig:long_endgame1_mcts}). While playing A is a perfectly fine move, as it forms a bridge with cells B and C, it unnecessarily extends the length of the game.

\begin{figure}[h!]
    \centering
    \includegraphics[width=0.35\linewidth]{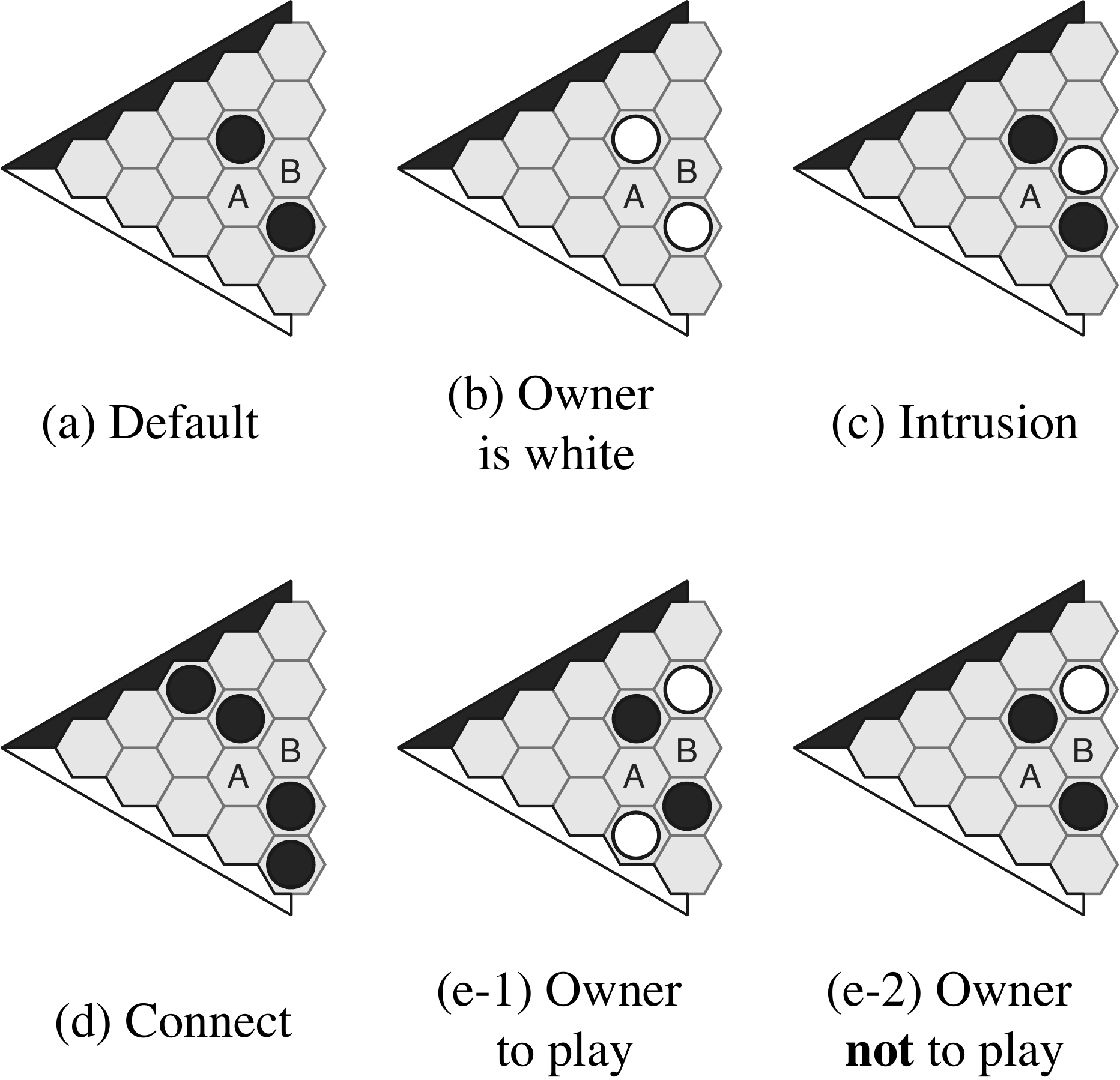}
\caption{\textbf{Conditions.} We generate boards from all combinations of these conditions. Only ``connect'' (d) makes a significant impact on the results.}
         \label{app:fig:conditions}
     \end{figure}

\begin{figure}[h!]
    \centering
      \raisebox{-0.5\height}{    \includegraphics[width=0.3\linewidth]{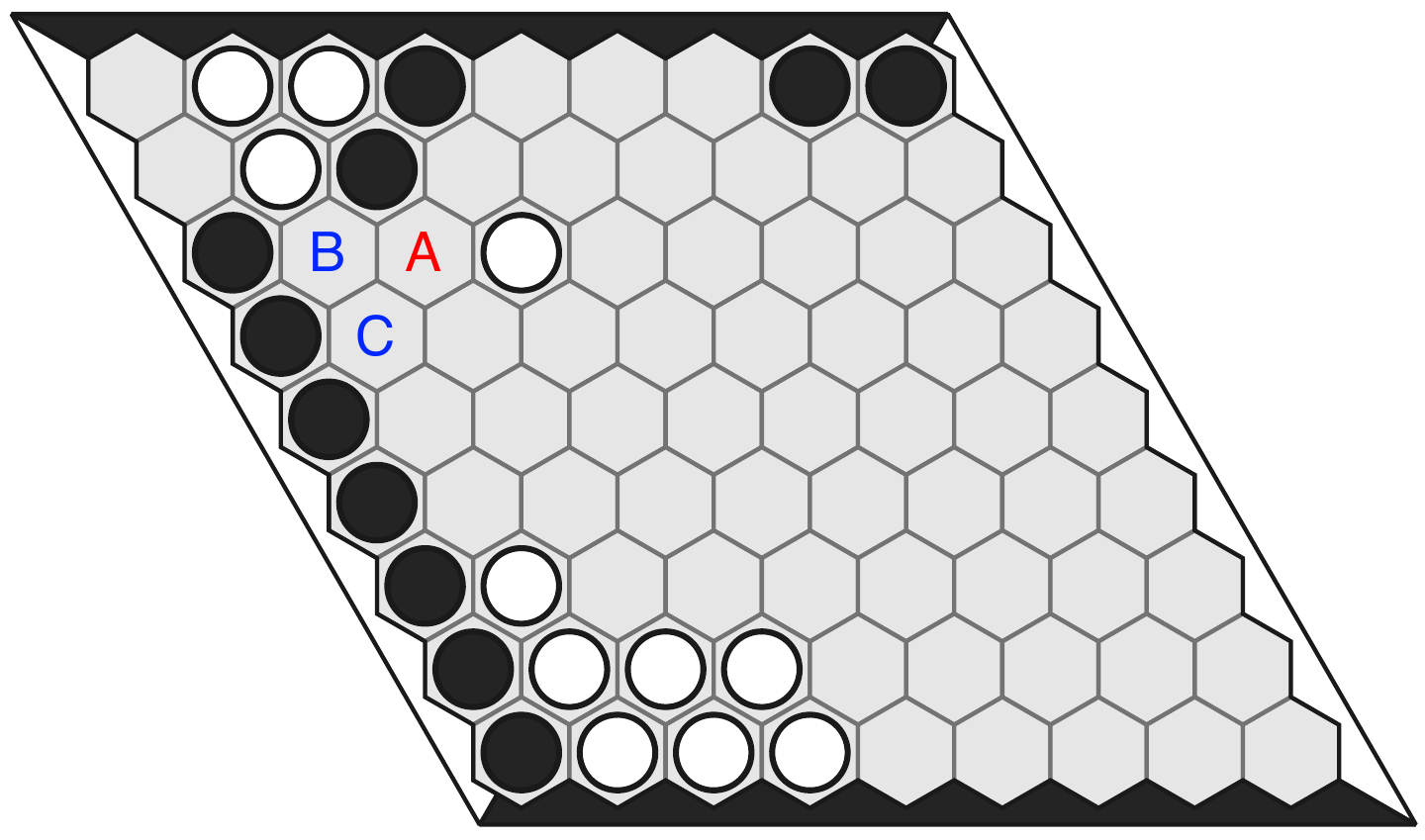}  }
  \hspace*{.01in}
  \raisebox{-0.5\height}{   \includegraphics[width=0.2\linewidth]{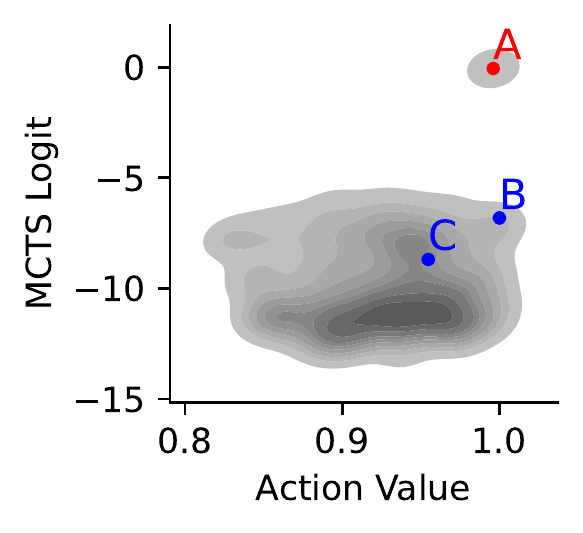}}
\caption{\textbf{Example end game board to test efficiency of last moves.} Black is close to winning the game and must decide its next move. Selecting the B ends the game, but AlphaZero instead selects A. While selecting A gives AlphaZero the ability to win the game on the next round, it unnecessarily extends the game. Right: While action B leads to a value of 1, action A also has a high value.
See Appendix \cref{app:fig:long_endgame1_mcts} for the per action MCTS logits.}
         \label{app:fig:long_endgame1}
     \end{figure}
     
\begin{figure}[h!]
    \centering
      \raisebox{-0.5\height}{    \includegraphics[width=0.3\linewidth]{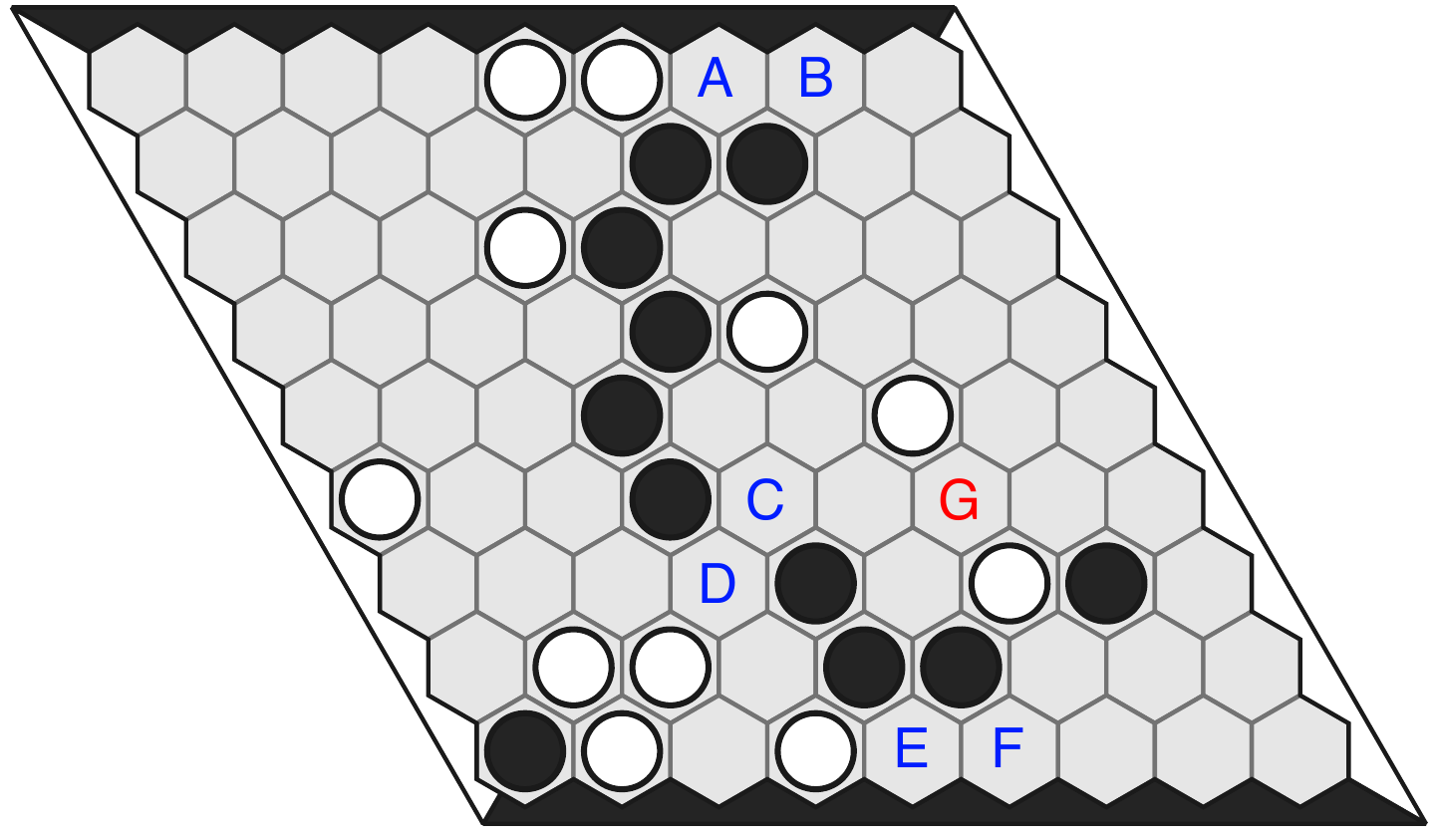}  }
  \hspace*{.01in}
  \raisebox{-0.5\height}{   \includegraphics[width=0.2\linewidth]{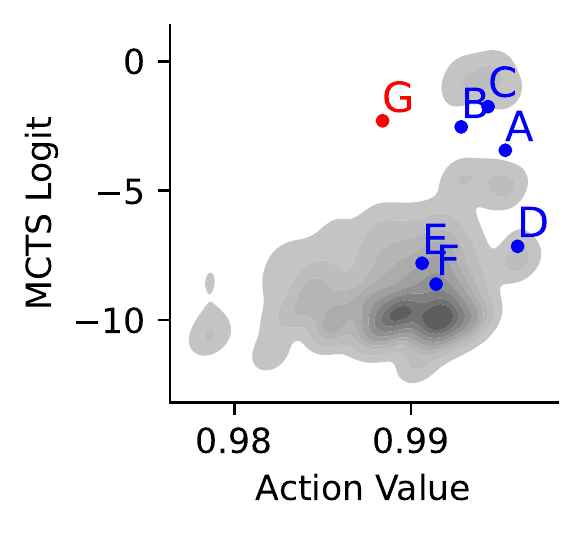}}
\caption{\textbf{Additional example endgame from selfplay.} There are six actions that move towards ending the game for black, (A-F). Sometimes, AZ instead plays G, which doesn't meaningful impact the game state. Right: Most the action values are above 0.98. Thus, the values have less impact on the MCTS logits. In 49 of 100 selfplay playouts continued from this position, AZ takes more actions than necessary. See App \cref{app:fig:long_endgame1_mcts} for the per action MCTS logits.}
\label{app:fig:long_endgame2}
\end{figure}

\begin{figure}[h]
    \centering
    \includegraphics[width=.9\textwidth]{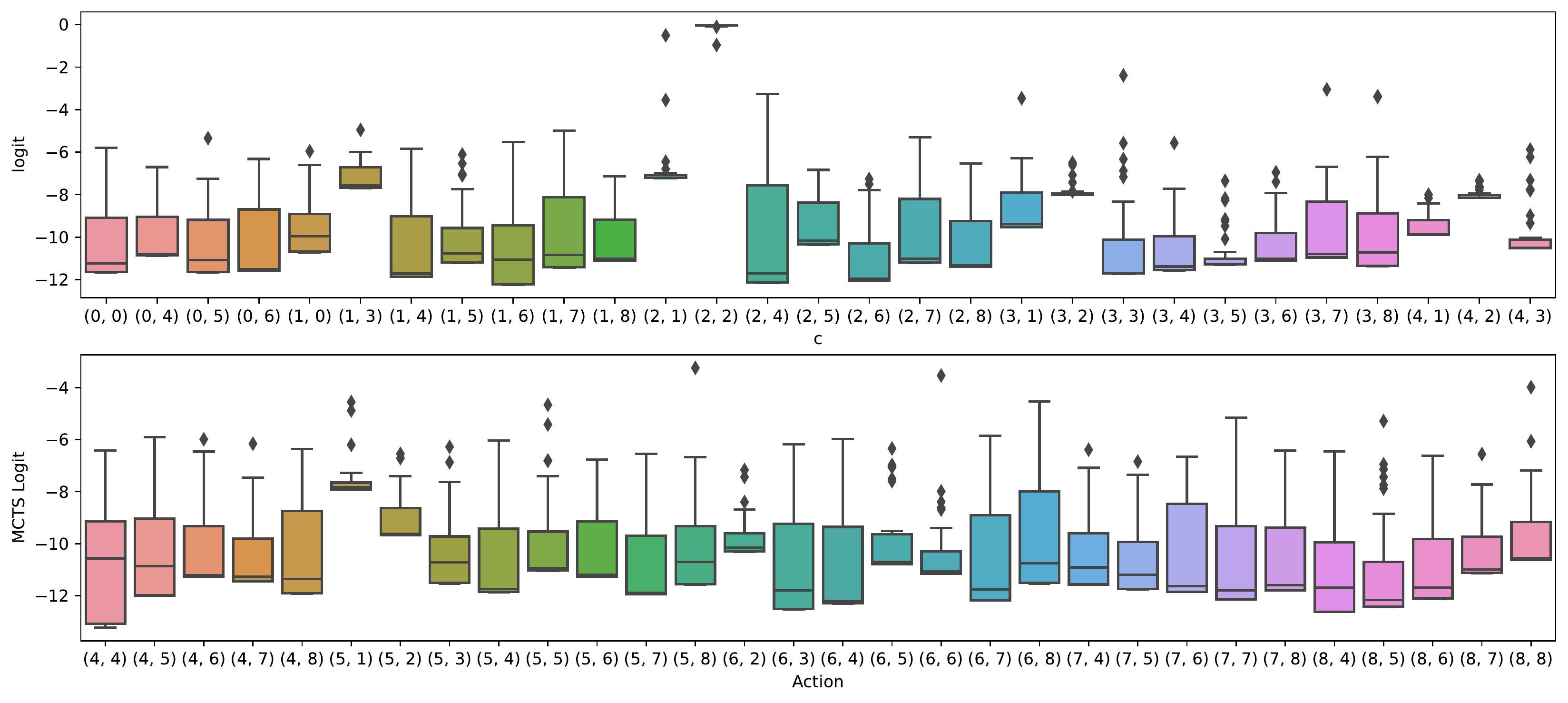}
    \caption{\textbf{Boxplot of MCTS logit values of example board presented in Figure \ref{app:fig:long_endgame1}.} While move 19 has high probability of being selected as the action for this board, AlphaZero places higher probability on move 20.  Move 19 results in an immediate win, while move 20 results in a win in the following round of gameplay.}
    \label{app:fig:long_endgame1_mcts}
\end{figure}

When AZ's value estimates are all high, it often takes actions that unnecessarily extend the length of the game. In \cref{app:fig:long_endgame2}, we present an additional example of AZ's behavior during an endgame from a selfplay rollout. AZ has three bridges (Figure \ref{app:fig:concepts}a) that will allow it to win the game: one with the top edge, one in the middle of the board, and one with the bottom edge. Thus, AZ is virtually connected to both edges, and guaranteed a win. However, all possible moves for black result in value estimates greater than 0.99, leading AlphaZero's MCTS to not produce significantly higher logit scores for the moves that lead to the win with fewest moves. The flatness of these value estimates suggests a lack of distinction between efficient and inefficient paths to victory.

This phenomena may have also occurred in AlphaGo's match against Lee Sedol. In game two, AlphaGo, the predecessor to AlphaZero (AZ), played a move the commentators deemed ``slack'' \citep{omerod2016}. This occurred towards the end of the game (move 167 out of 211). Given that AlphaGo (and AZ) only consider the probability of winning, not the margin (nor the timeframe), again, it seems that it will sometimes waste moves when it it believes the game to be won. So, was move 167 a mistake? Was playing A in \cref{app:fig:long_endgame1} or G in \cref{app:fig:long_endgame2} a mistake? They are not (necessarily) mistakes, because AZ can still win. AZ may understand it can use the relevant concepts to win the game, and may eventually do so. This has ramifications for how we test AZ's behavioral understanding of concepts. To  determine if AZ uses a concept, we present it with a situation where understanding the concept and using it is the only way to win.

\begin{figure}[h]
    \centering
    \includegraphics[width=.9\textwidth]{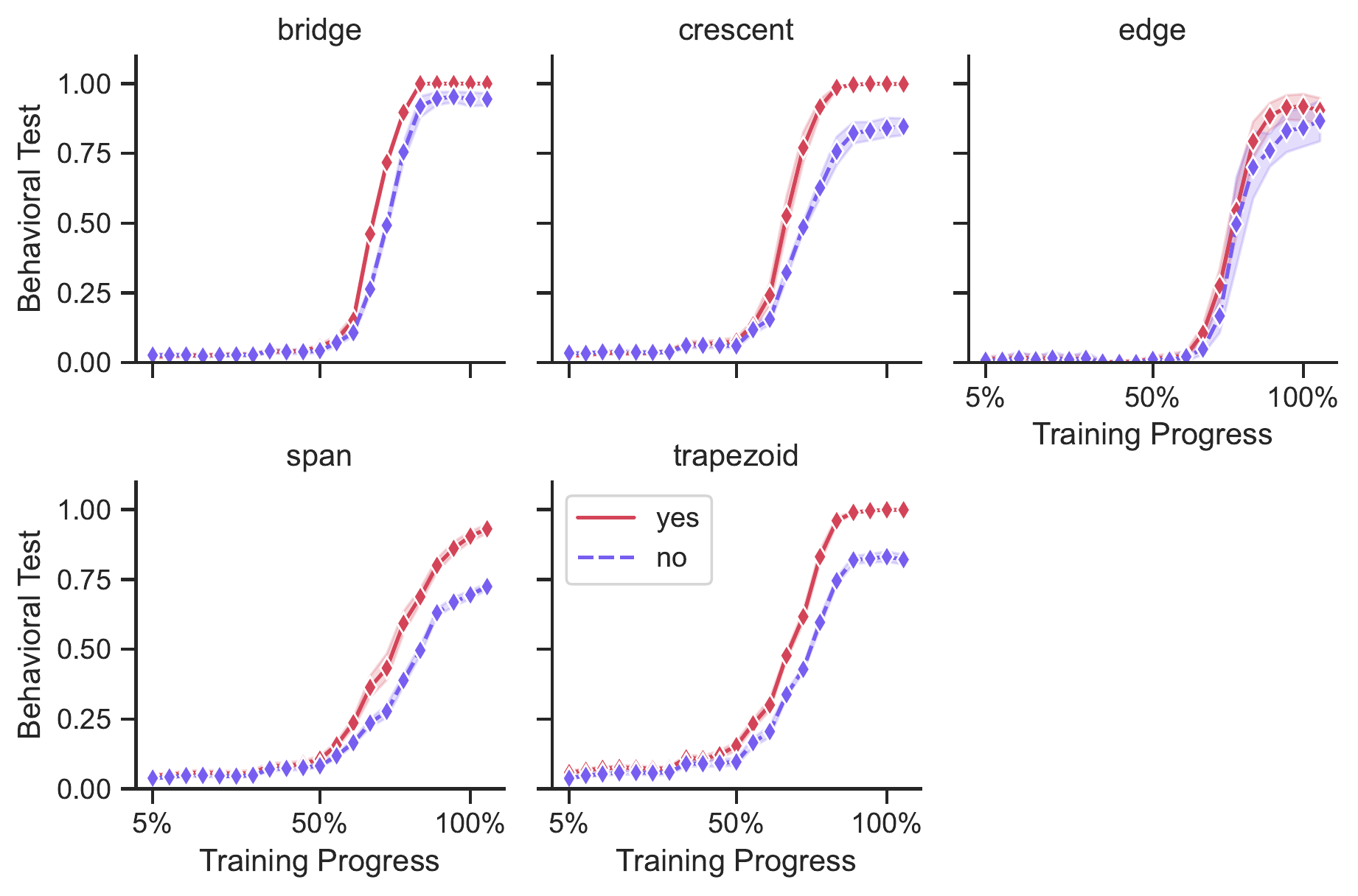}
    \caption{\textbf{Comparing behavioral results for connected vs. not connected defenders.} The passing results are lower than the connected instance because the model does not always choose the fastest route to victory. }
    \label{app:fig:connectedornot}
\end{figure}

Because AZ does not prefer faster routes, as described in the main paper, we test the agents in forced situations. In \cref{app:fig:connectedornot}, not connecting the defender -- i.e., testing the agent in unforced situations, the passing rate of the behavioral test is lower. This is because the agent takes a longer route to victory, not necessarily because the agent does not evaluate the concept/correct-behavior as winning. 


\clearpage
\section{Additional Results}\label{app:sec:results}
\begin{figure}
    \centering
    \includegraphics[width=.99\textwidth]{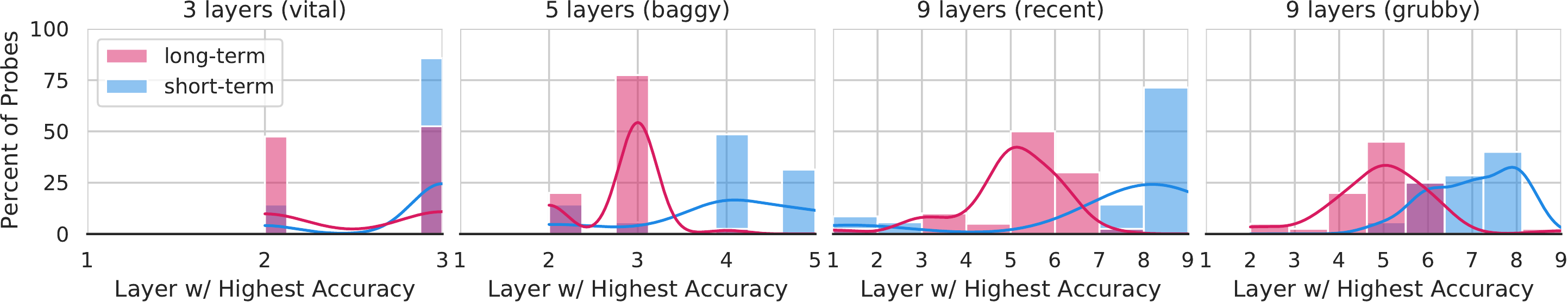}
    \includegraphics[width=.99\textwidth]{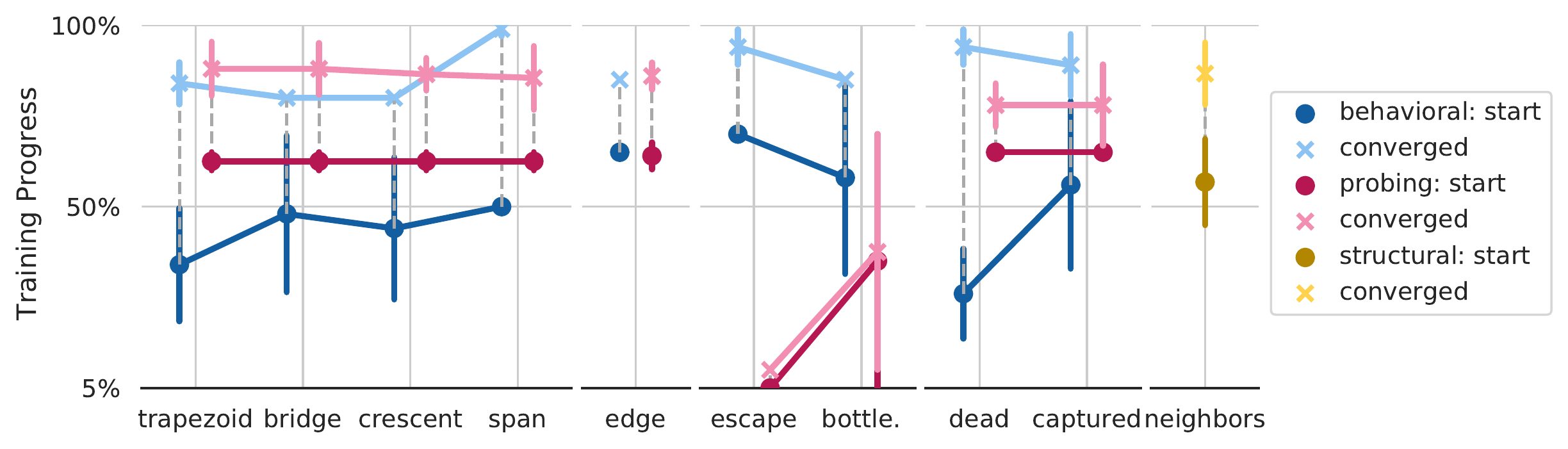}
    \includegraphics[width=.99\textwidth]{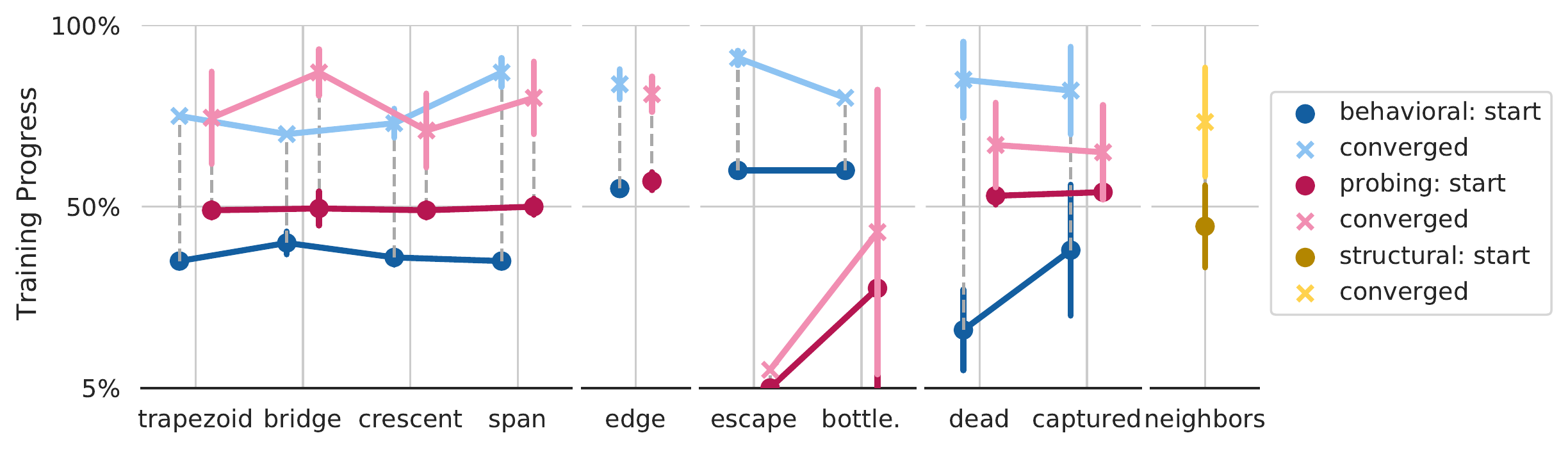}
    \includegraphics[width=.99\textwidth]{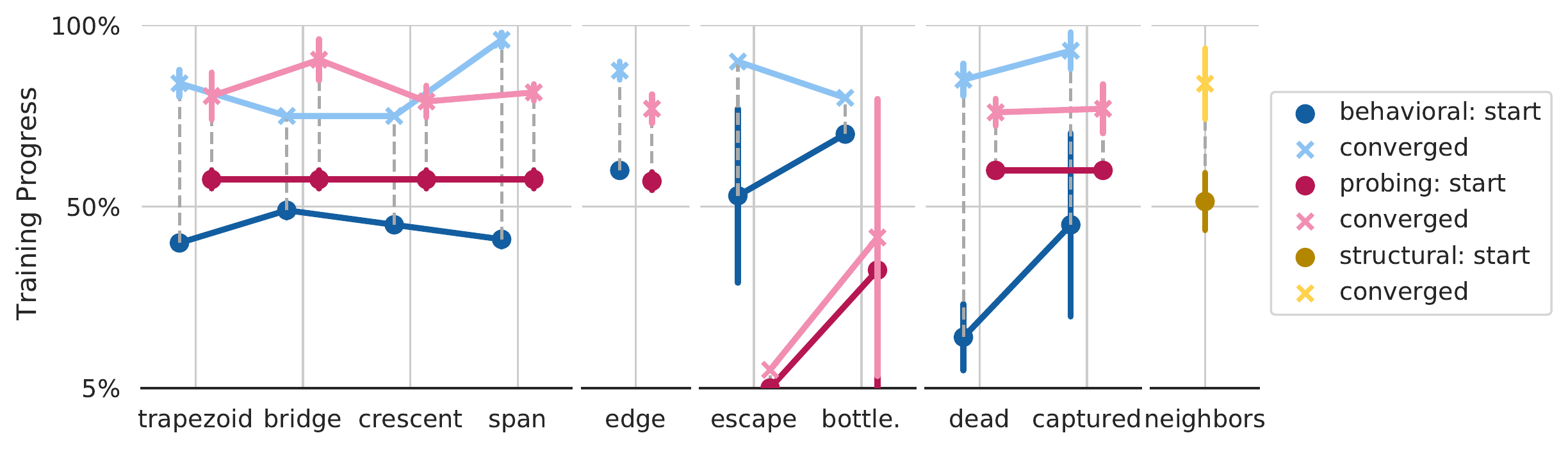}
    \includegraphics[width=.99\textwidth]{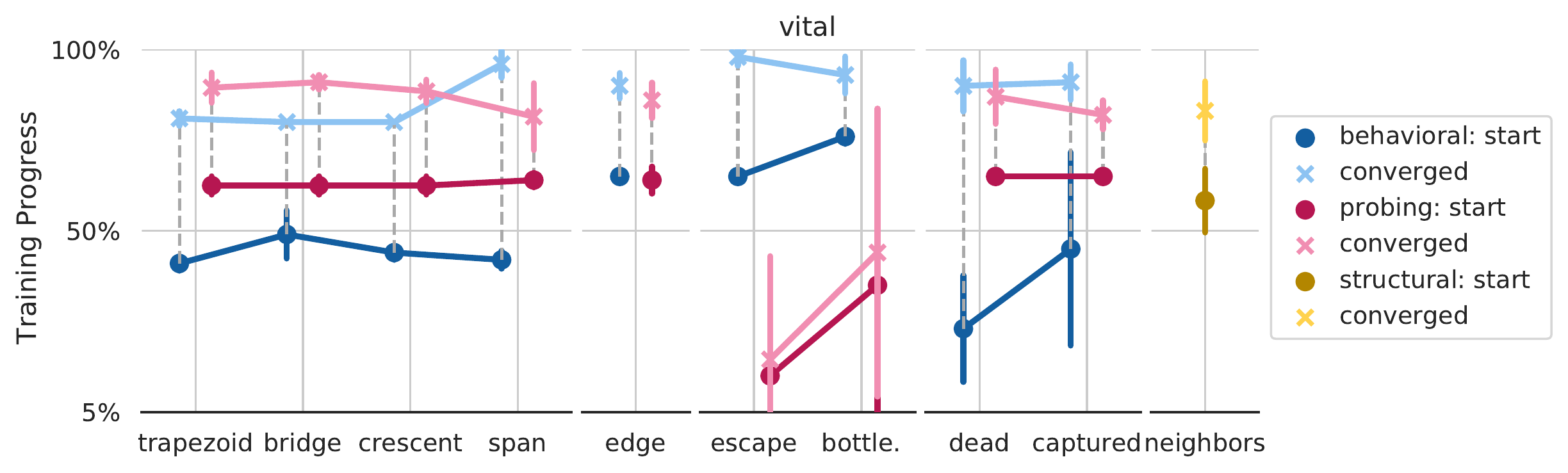}
    \caption{\textbf{Key figures from the main paper replicated across architectures. }}
    \label{app:fig:cross-arch}
\end{figure}

\begin{figure}
    \centering
    \includegraphics[width=.55\linewidth]{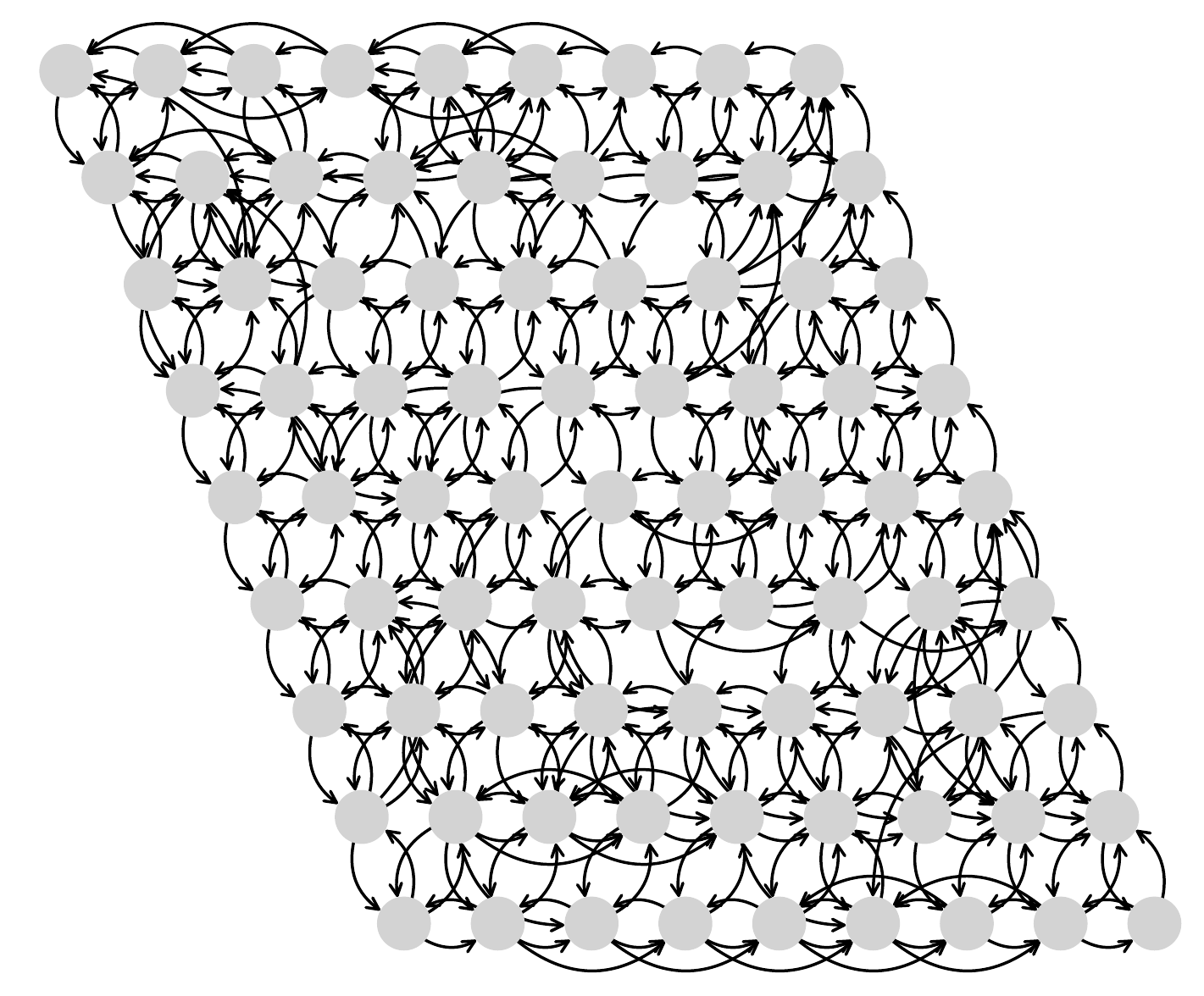}
    \includegraphics[width=.40\textwidth]{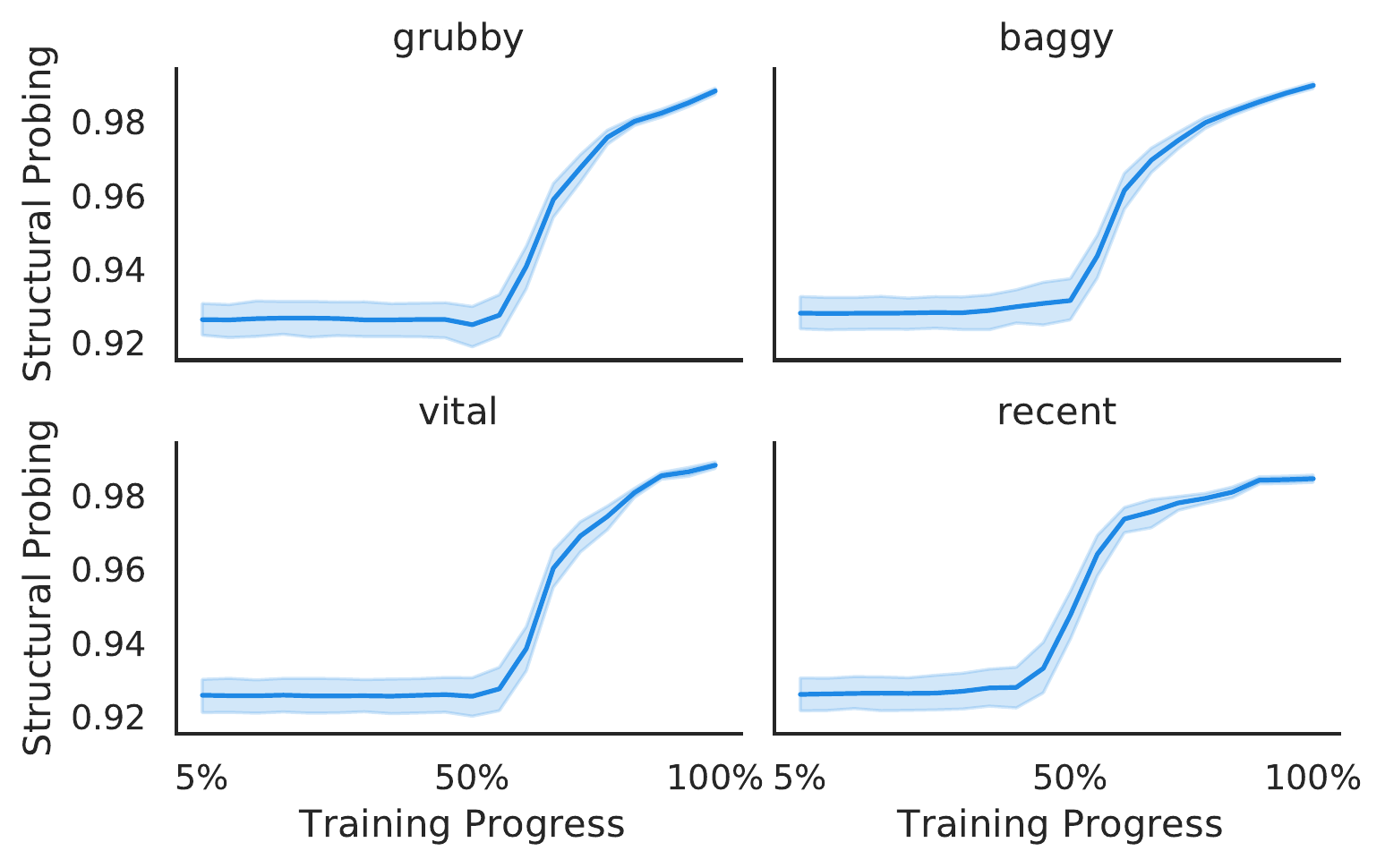}
    \includegraphics[width=\textwidth]{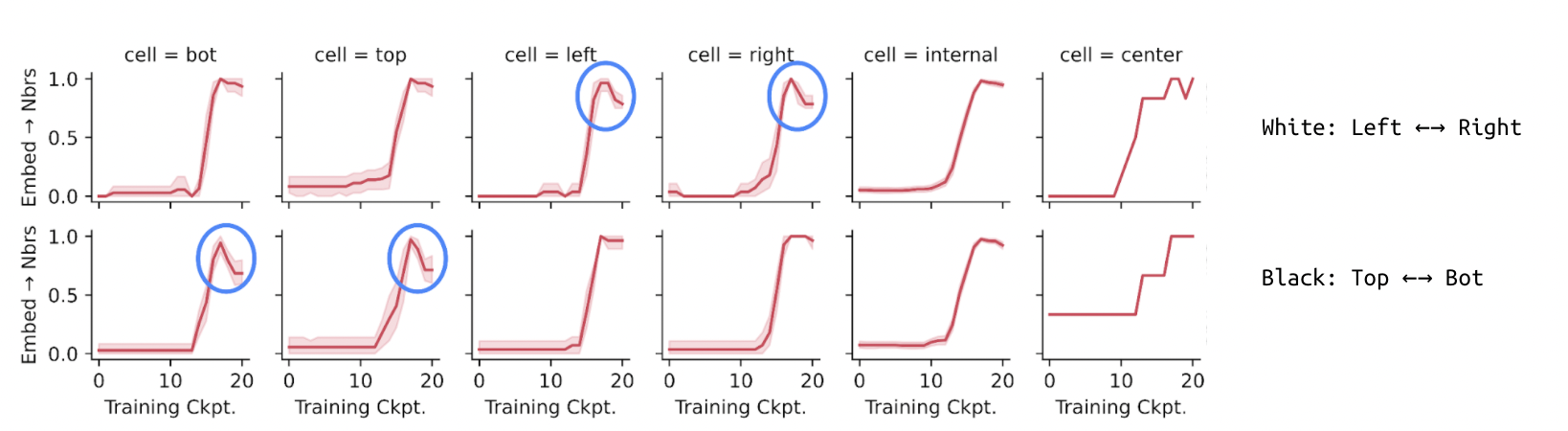}
    \caption{\textbf{Left:} \textbf{Dot-product scores between cell embeddings recover board structure, with modifications on the upper and lower edges.} Each grey circle corresponds to the same cell on the Hexboard; the black arrows correspond to its nearest neighbors according to dot-product scores between cells' embeddings. This diagram shows the neighborhood structure for the final checkpoint in training. See the evolution of the board structure here: \url{https://drive.google.com/file/d/1UV6mhnJ_FJOP3fEiHITp0bUDC7lUMJx9/view?usp=sharing}.
    \textbf{Right:} \textbf{Cell embeddings encode the board structure of the game only 50$\%$ of the way through training.} We measure how well dot-product scores between cells align with the ground truth cell distance using a ranking metric Normalized Discounted Cumulative Gain (NDCG). \textbf{Bottom:} This chart partitions the scoring of the neighbors by cell location (bot, top, left, right, internal, center), as well as, by player. Interestingly, there is a slight deviation from the increase in neighorhood structure aligning with the underlying grid for a player's own edge cells. This appears to be from the edge pieces ``connecting'' to each other at distances of two--the extra loops along the bottom and top rows.}
    \label{app:fig:board_structure}
\end{figure}

\begin{figure}[h!]
        \centering
        \includegraphics[width=.485\linewidth]{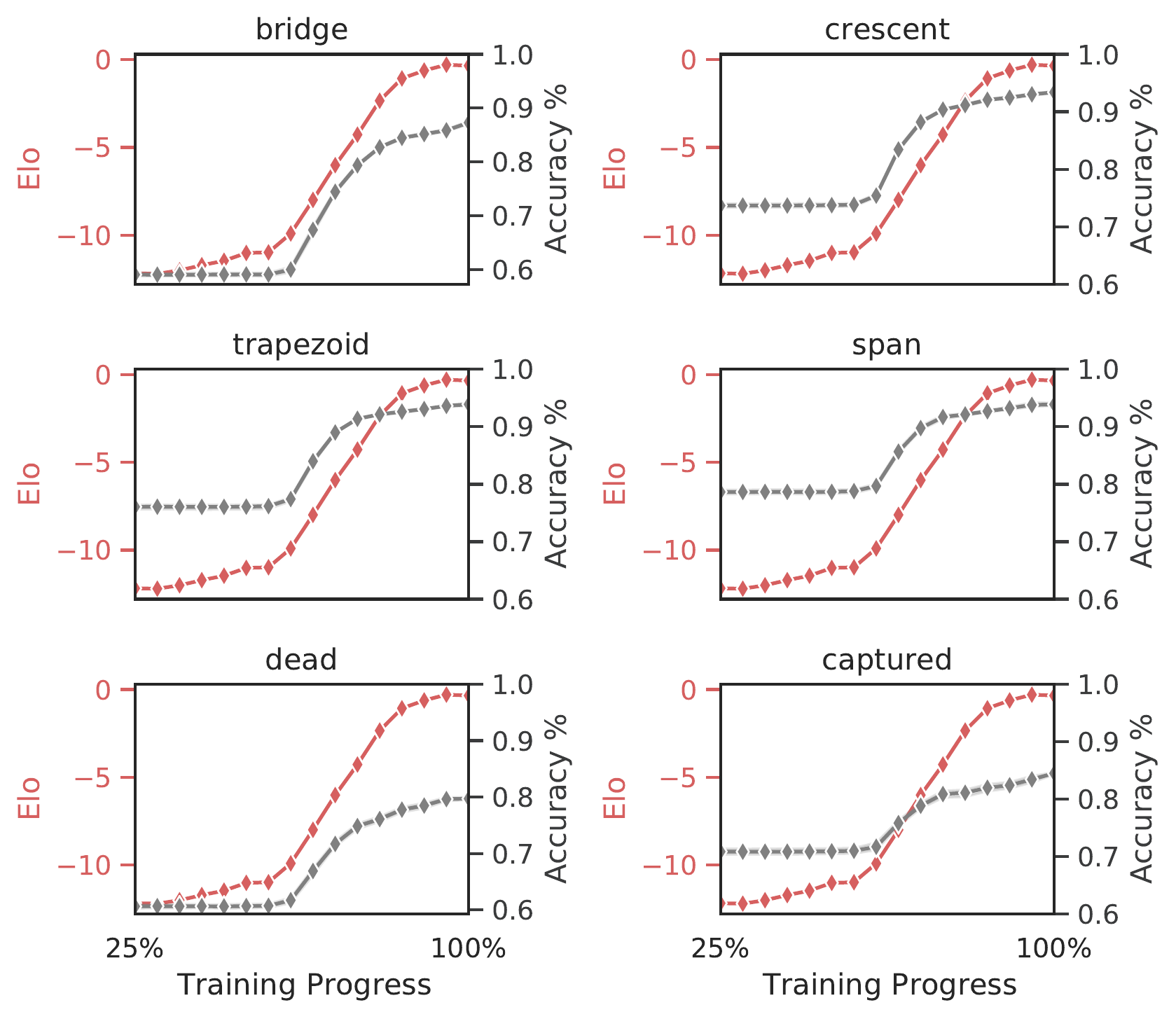}
        \includegraphics[width=.485\linewidth]{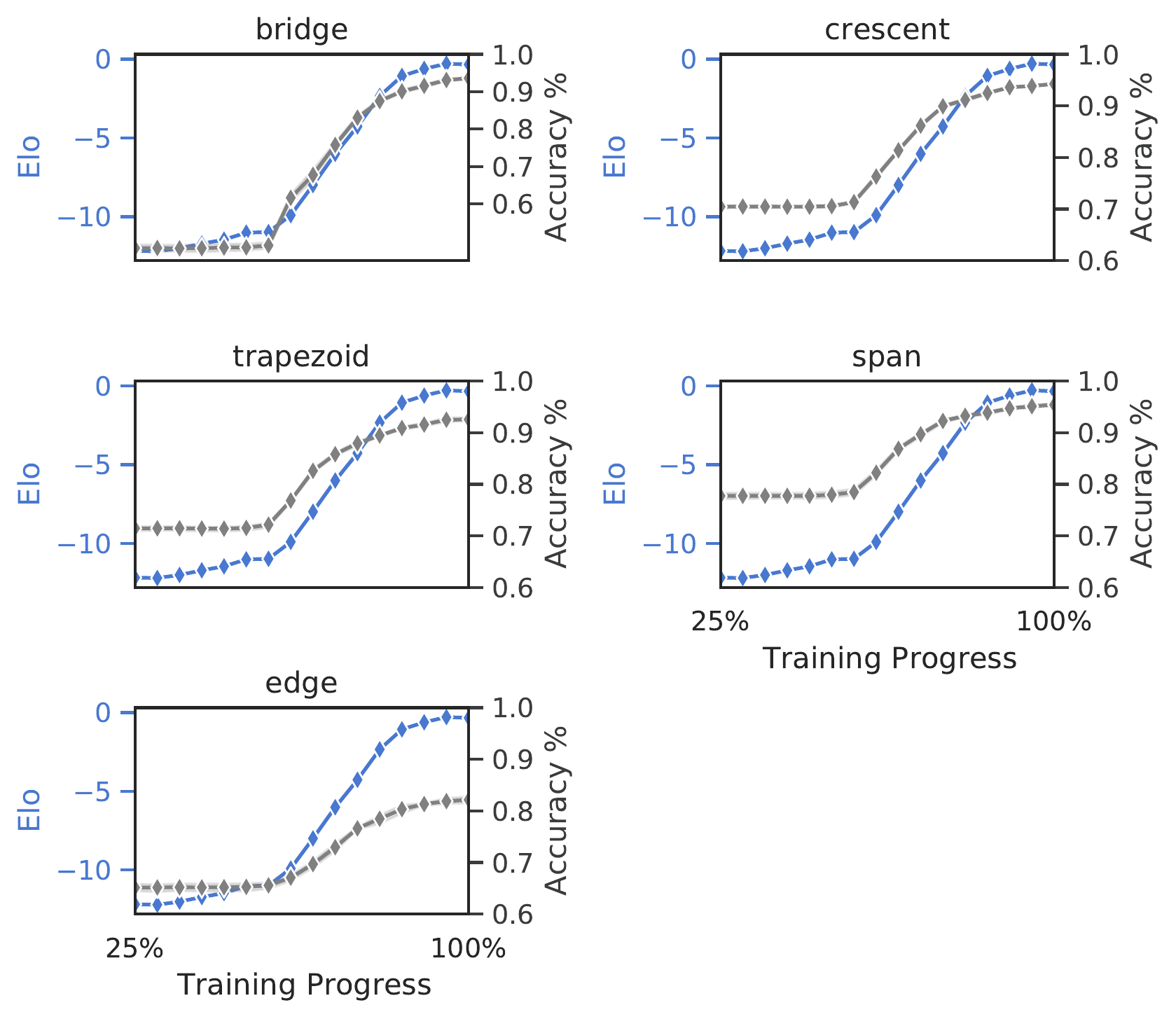}
    \caption{\textbf{Improvements in gameplay ability, as measured by Elo, coincide with improvements in concept recognition, as measured by the test accuracy of the linear probe on the highest performing layer.}}
             \label{fig:elo}
    \end{figure}

\begin{figure}[h!]
        \centering
        \includegraphics[width=0.485\linewidth]{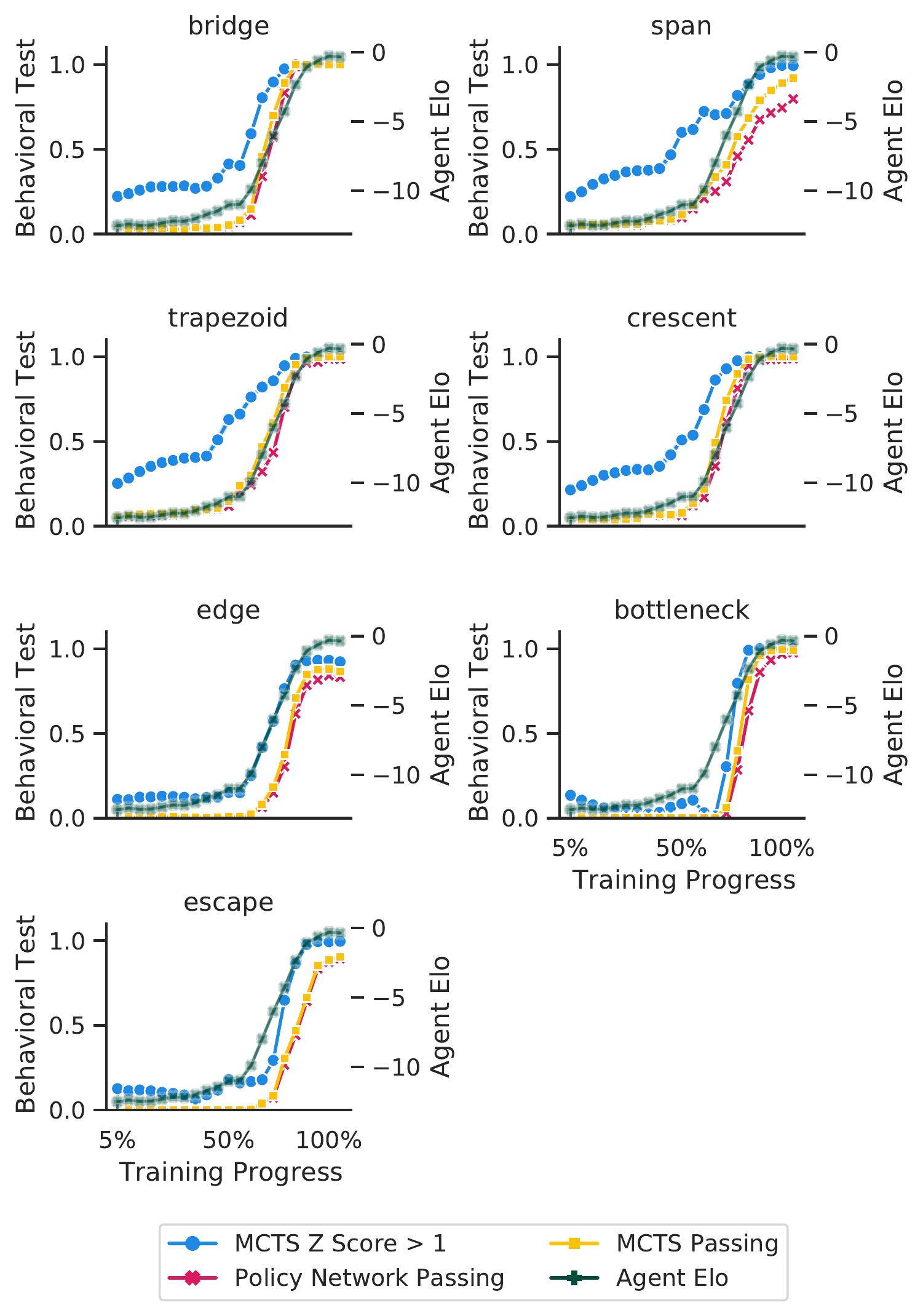}
        \includegraphics[width=0.485\linewidth]{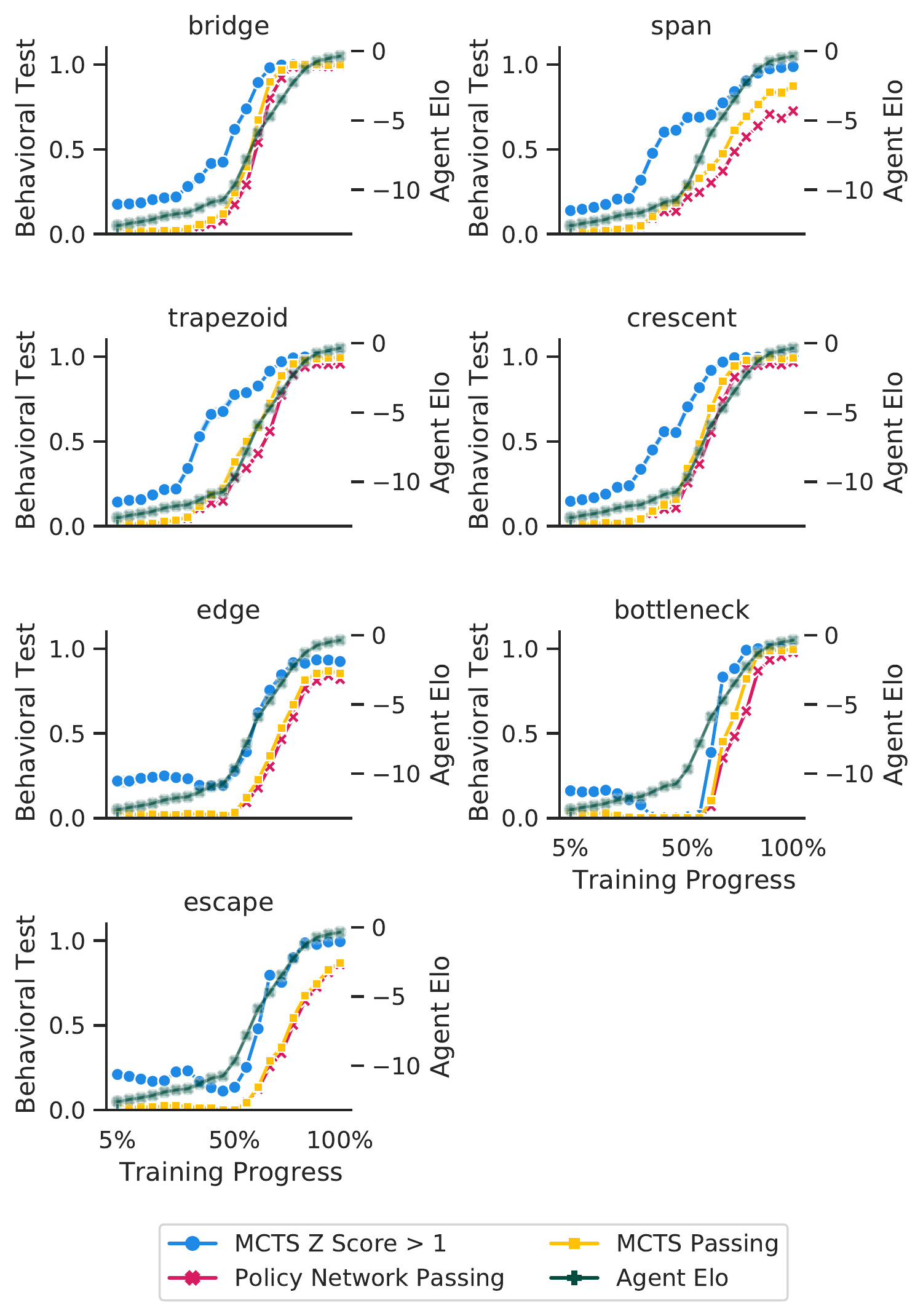}
    \caption{\textbf{AlphaZero learns to use the positive concepts; Model Code = (left) \texttt{grubby}, (right) \texttt{recent}}. See \cref{app:tbl:agents} for architecture details of the models. At each checkpoint, we present AlphaZero with a set of example boards that test its ability to utilize each concept. MCTS (yellow) and the deep policy network (red) select actions that pass our behavioral tests with increasing frequency throughout training. We additionally report the rate at which the action that passes our behavioral test is one standard deviation above the mean (z score $> 1$). The Agent Elo (dark green) measures AZ's general gameplaying ability; it increases as AlphaZero starts to use the concepts.
    }
             \label{app:fig:positive_1}
         \end{figure}

\begin{figure}[h!]
        \centering
        \includegraphics[width=0.485\linewidth]{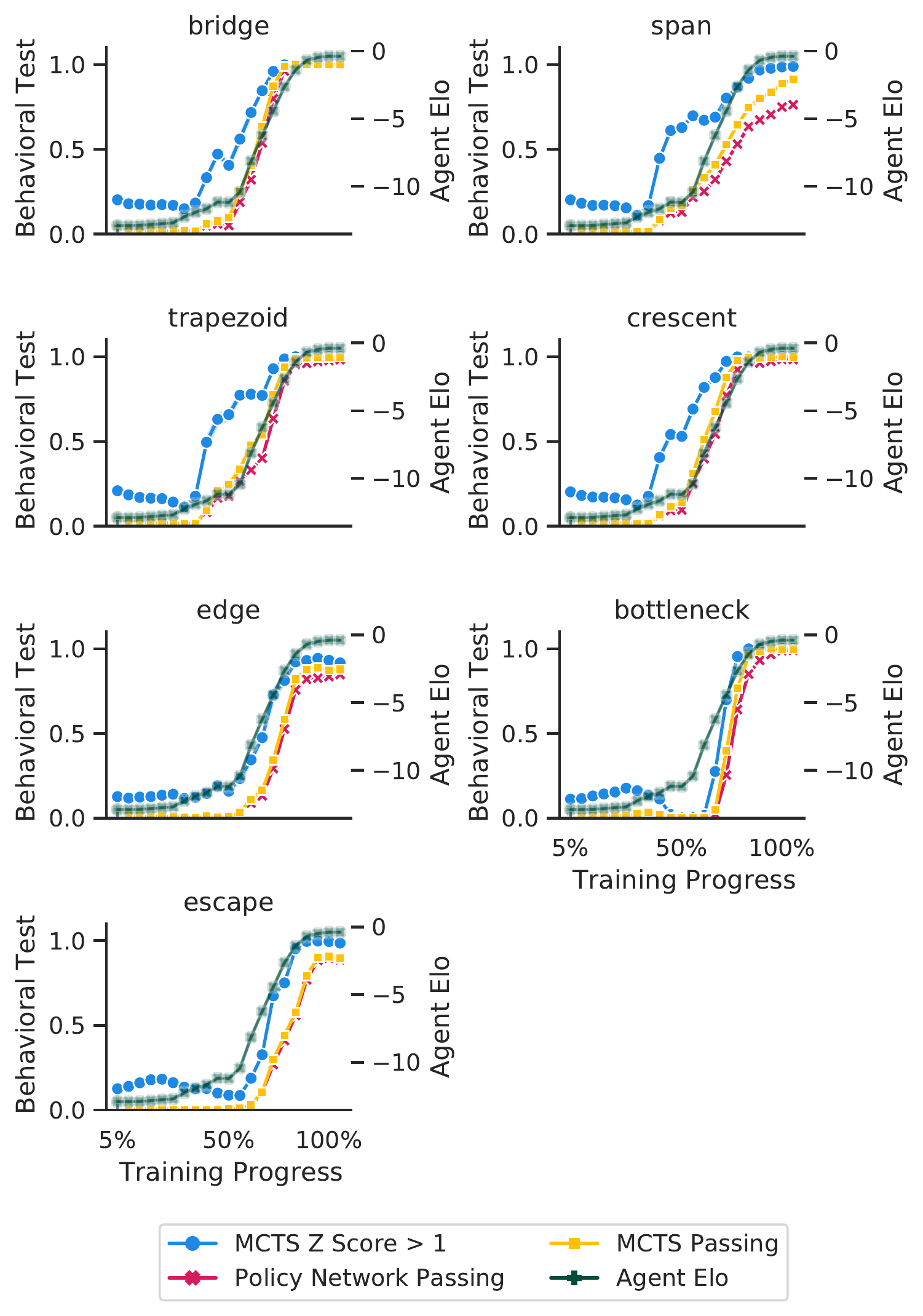}
        \includegraphics[width=0.485\linewidth]{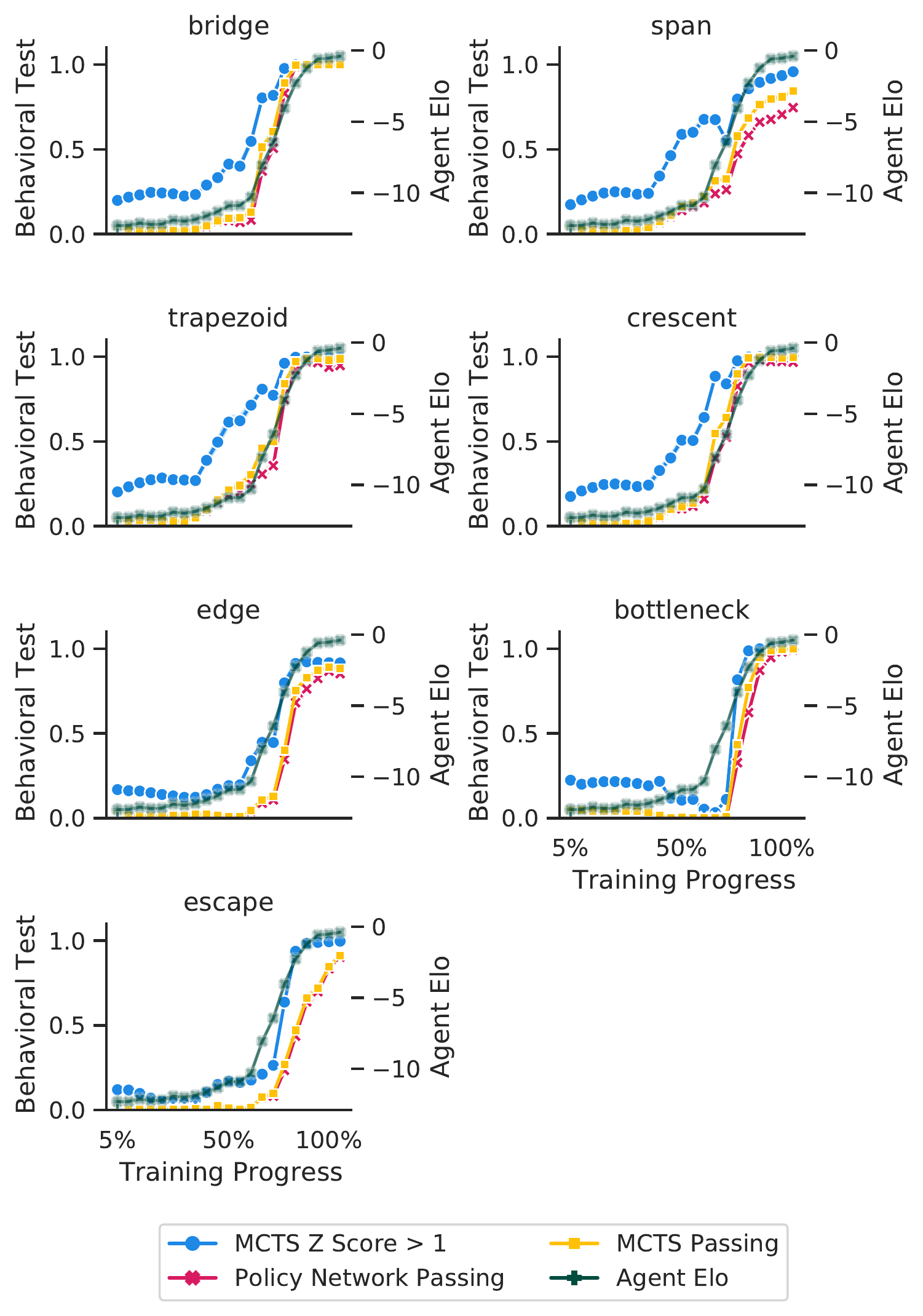}
    \caption{\textbf{AlphaZero learns to use the positive concepts; Model Code = (left) \texttt{baggy}, (right) \texttt{vital}}. See \cref{app:tbl:agents} for architecture details of the models. At each checkpoint, we present AlphaZero with a set of example boards that test its ability to utilize each concept. MCTS (yellow) and the deep policy network (red) select actions that pass our behavioral tests with increasing frequency throughout training. We additionally report the rate at which the action that passes our behavioral test is one standard deviation above the mean (z score $> 1$). The Agent Elo (dark green) measures AZ's general gameplaying ability; it increases as AlphaZero starts to use the concepts.
    }
             \label{app:fig:positive_2}
         \end{figure}

\begin{figure}[h!]
        \centering
        \includegraphics[width=0.45\linewidth]{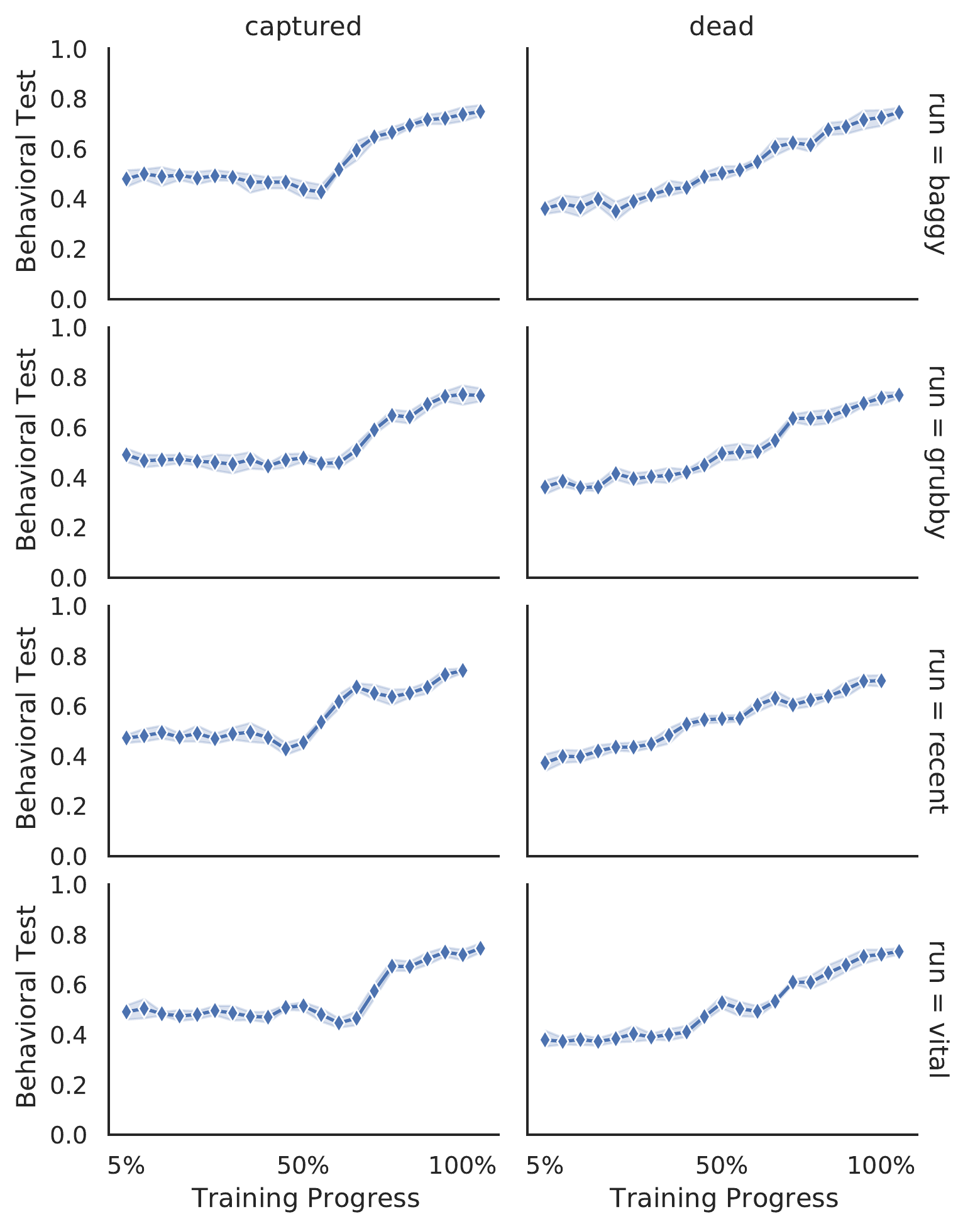}
    \caption{\textbf{AlphaZero does not fully ``use'' the negative concepts.}
    At the end of training, it plays moves in 25\% of our behavioral tests that cannot impact the outcome of the game \citep{bjornsson2006dead}. To pass these behavioral tests, AlphaZero must \textit{avoid} playing cells on the board associated with the dead and captured concepts throughout a full selfplay rollout.
    }
             \label{app:fig:negative}
         \end{figure}

\cref{app:fig:cross-arch} reports key figures from the main paper replicated across architectures. \cref{app:fig:board_structure} shows the same board structure from the main body of the paper, along with some of the quantative measures of the board structure. \cref{fig:elo}, \cref{app:fig:positive_1}, \cref{app:fig:positive_2}, \cref{app:fig:negative} report the behavioral test scores (as well as ``elo'').

\end{document}